\documentclass[10pt,twocolumn,letterpaper]{article}

\usepackage[pagenumbers]{cvpr} % To force page numbers, e.g. for an arXiv version

\usepackage{graphicx}
\usepackage{amsmath}
\usepackage{amssymb}
\usepackage{booktabs}
\usepackage{multirow}
\usepackage{array}
\usepackage{float}
\usepackage{url}
\usepackage[accsupp]{axessibility}  % Improves PDF readability for those with disabilities.

\usepackage[pagebackref,breaklinks,colorlinks]{hyperref} 
\usepackage[capitalize]{cleveref}
\crefname{section}{Sec.}{Secs.}
\Crefname{section}{Section}{Sections}
\Crefname{table}{Table}{Tables}
\crefname{table}{Tab.}{Tabs.}

\def\cvprPaperID{5128} % *** Enter the CVPR Paper ID here
\def\confName{CVPR}
\def\confYear{2022}

\begin{document}

\title{Make It Move: Controllable Image-to-Video Generation with Text Descriptions}

\author{Yaosi Hu\textsuperscript{1}\thanks{This work was done while Yaosi Hu was an intern at MSRA.} ~~~  
Chong Luo\textsuperscript{2}  ~~~
Zhenzhong Chen\textsuperscript{1} ~~~
\\
Wuhan University\textsuperscript{1} ~~~
Microsoft Research Asia\textsuperscript{2}\\
{\tt\small ys\_hu@whu.edu.cn ~~~ 
cluo@microsoft.com ~~~
zzchen@whu.edu.cn}
}
\maketitle

%%%%%%%%% ABSTRACT
\begin{abstract}
   Generating controllable videos conforming to user intentions is an appealing yet challenging topic in computer vision. To enable maneuverable control in line with user intentions, a novel video generation task, named Text-Image-to-Video generation (TI2V), is proposed. With both controllable appearance and motion, TI2V aims at generating videos from a static image and a text description. 
   The key challenges of TI2V task lie both in aligning appearance and motion from different modalities, and in handling uncertainty in text descriptions. To address these challenges, we propose a Motion Anchor-based video GEnerator (MAGE) with an innovative motion anchor (MA) structure to store appearance-motion aligned representation. To model the uncertainty and increase the diversity, it further allows the injection of explicit condition and implicit randomness. Through three-dimensional axial transformers, MA is interacted with given image to generate next frames recursively with satisfying controllability and diversity.
   Accompanying the new task, we build two new video-text paired datasets based on MNIST and CATER for evaluation. Experiments conducted on these datasets verify the effectiveness of MAGE and show appealing potentials of TI2V task. Code and datasets are released at \url{https://github.com/Youncy-Hu/MAGE}.
\end{abstract} 
%%%%%%%%% BODY TEXT

\vspace{-1.5em}
\section{Introduction}
\label{sec:intro}
Video generation has undergone revolutionary changes and has made great progress in recent years. Early research of unconditional video generation \cite{NIPS2016_04025959, Saito_2017_ICCV, Tulyakov_2018_CVPR} focused on how to generate a video from noise or a latent vector from an aligned latent space. Recently, more emphases have been put on controllable video generation \cite{Hao_2018_CVPR, nips2019vid2vid, Blattmann_2021_CVPR}, which allows users to express their intentions about how the scene or the objects look like (appearance information) or how the objects move (motion information).
Controllable video generation has many potential applications, including facilitating designers in artistic creation and assisting machine learning practitioners for data augmentation. 

\begin{figure}[htbp]
    \centering
    \includegraphics[width=0.45\textwidth]{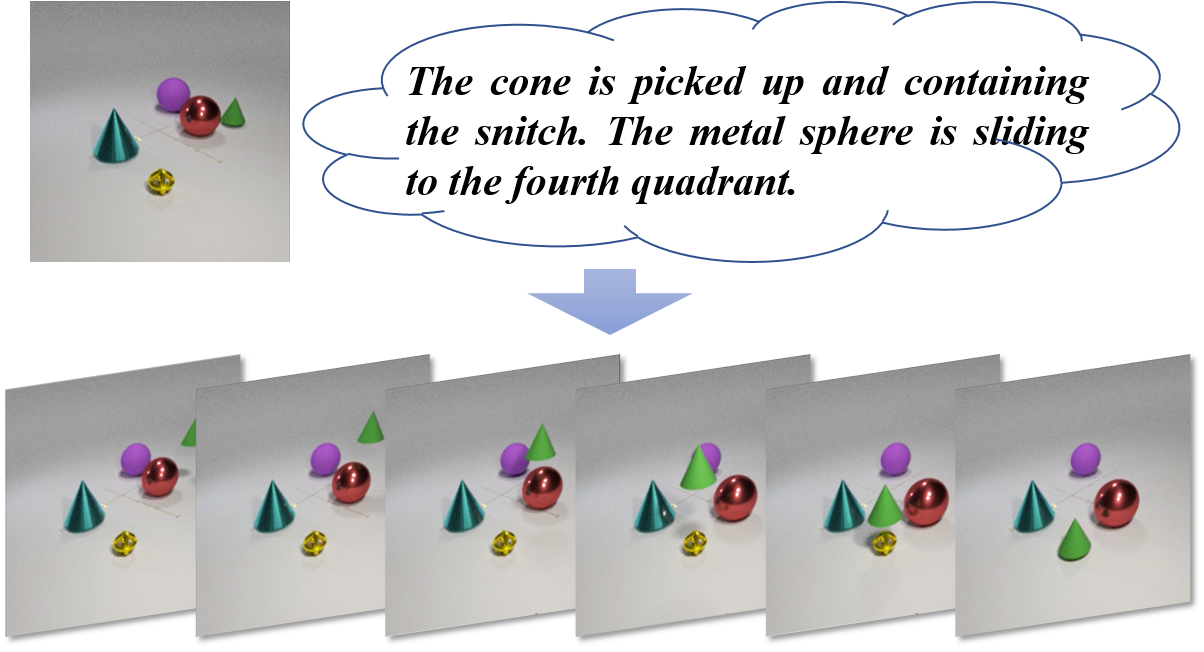}
    \caption{An illustration of the proposed TI2V task. An image and a detailed text description provide the appearance and motion information for video generation, respectively.}
    \label{fig:example}
\vspace{-1.0em}
\end{figure}

Existing controllable video generation tasks can be grouped into three categories, namely Image-to-Video generation (I2V), Video-to-Video generation (V2V), and Text-to-Video generation (T2V). These tasks provide different ways for users to inject the appearance and the motion information and therefore have different levels of control over these two factors. 
I2V and V2V have strong control over the appearance of generated video, as separate images are usually provided to set the scene. As for the motion, I2V shows limited controllability since the task is defined to accept only coarse-grained motion clues, such as predefined action labels or directions \cite{Blattmann_2021_CVPR}. In contrast, V2V can generate videos with highly controllable motion because detailed motion guidance, such as trajectories \cite{Hao_2018_CVPR} or action sequences \cite{Chan_2019_ICCV, Menapace_2021_CVPR}, are provided in the form of input video. But one drawback of V2V in practical use is that such motion guidance is hard to be obtained. Among all the three tasks, T2V has the weakest control over the generated video. Users provide both appearance and motion information through text, which is imprecise and sometimes ambiguous. Nevertheless, text description of motion is more in line with human habits \cite{wu2021godiva} and leaves a lot of room for creation and imagination in video generation. 

In this paper, we introduce a novel video generation task, named Text-Image-to-Video generation (TI2V). It provides a natural way for users to express their intentions, using a single static image to set the scene and a natural text description to provide motion.
TI2V is a more difficult task than I2V or T2V. It not only requires the separate understanding of text and image, but also needs to align visual objects with corresponding text descriptions, and then transform the implied object motion to an explicit video. 
We aim to achieve two goals in the TI2V task: \expandafter{\romannumeral1}) \textbf{Controllable}. Under the constraints of image and text, the generated video should have visually consistent appearance set by the given image and semantically aligned motion as described in the text. \expandafter{\romannumeral2}) \textbf{Diverse}. This goal resolves ambiguity and brings creativity, which are important and appealing features for video generation. In the example given in Fig.\ref{fig:example}, the text description does not specify which cone it wants to pick up and which exact position in ``the fourth quadrant" it wants the metal sphere to slide to. Under such ``constrained randomness", we want to produce videos that match the description but are also diverse. 

We design an auto-regressive framework, named MAGE, to address the TI2V task. A VQ-VAE encoder-decoder architecture is adopted for efficient visual token representation. The key challenge is how to merge the text-described motion into visual features to generate a controllable and diverse video. 
To achieve the controllable goal, we propose a spatially aligned Motion Anchor (MA) to integrate the appearance and motion information through the cross-attention operation in the common latent space for image and text. Each position in the MA stores all necessary motion information of the corresponding region for video generation. We further introduce explicit condition and implicit randomness into MA. The explicit condition provides additional constraint from an explicit input (e.g., speed) to improve both controllability and diversity, while the implicit randomness brings in uncertainty in the data distribution, allowing the model to generate diverse videos in a stochastic way. 
In the proposed MAGE framework, we adopt axial transformer to inject and fuse MA into visual tokens and generate videos in an auto-regressive manner.

To evaluate TI2V task and our generation model, appropriate paired video-text datasets are in need. Different from T2V task that often conducts experiments on action recognition datasets like KTH \cite{kth2004} or captioning datasets like MSR-VTT \cite{Xu_2016_CVPR} with action label or coarse-grained caption, TI2V focuses more on the maneuvering capability to image and requires fine-grained text description. Therefore, we propose two datasets with synthetic videos and fine-grained text descriptions based on MNIST \cite{MNIST1989} and CATER \cite{Girdhar2020CATER} for TI2V task. By controlling the uncertainty in descriptions, we can evaluate the performance of both deterministic and stochastic video generation.

The contributions of this paper are concluded as follows:
\begin{itemize}
   \item A novel Text-Image-to-Video generation task (TI2V) is introduced, aiming to generate visually consistent video from an image and a text description.
   \item A Motion Anchor-based video GEnerator (MAGE) is proposed to generate controllable and diverse videos. The core structure, motion anchor (MA), addresses the challenging matching problem between the appearance in the image and the motion clues in the text.
    \item Two video-text paired datasets modified from MNIST and CATER are built for the evaluation of TI2V task. Moreover, experiments conducted on these two datasets verify the effectiveness of MAGE.
\end{itemize}

\section{Related Work}
We only consider the work of video generation guided by human intention. 
In this context, human intention is mainly composed of the description of the scene (spatial information) and the description of the motion (temporal information). 
We classify related work into three categories, namely video-to-video (V2V) generation, image-to-video (I2V) generation, and text-to-video (T2V) generation, according to how human intention is expressed. 

Note that unconditional video generation \cite{NIPS2016_04025959, Saito_2017_ICCV, Tulyakov_2018_CVPR, Wang_2020_CVPR, wang2021inmodegan} is not discussed here, as these works generate videos from a random variable or a latent vector, and do not provide an interface for human to express their intention. 

%Text-to-Video generation (T2V) uses text to express both types of information. Video-to-Video generation (V2V) uses a video to express the motion information. Image-to-Video generation (I2V) uses an image to describe the scene and provides a limited set of motions for human to choose from.  
\subsection{Video-to-Video Generation}
There are two popular forms of V2V task, namely future video prediction and video-to-video synthesis. They retrieve or predict the desired motion from an input video. 

Future video prediction predicts future frames based on several past frames provided. The generator is required to retrieve past motion and predict the future \cite{Liang_2017_ICCV, Walker_2017_ICCV, pmlr-v80-wichers18a, Wu_2020_CVPR}. Due to the unpredictable nature of object motion, works in this setting are only used to predict very few future frames.  
% Yaosi: too many citations here, since they are not closely related, we only need to include a couple of high-impact ones \cite{pmlr-v70-kalchbrenner17a, 10.5555/2969442.2969560, Zhou_2015_ICCV, NIPS2016_d9d4f495, DBLP:journals/corr/MathieuCL15, Liang_2017_ICCV, Walker_2017_ICCV, Milbich_2017_ICCV, villegas2017decomposing, pmlr-v70-villegas17a, pmlr-v80-wichers18a, Cai_2018_ECCV, Wu_2020_CVPR, shrivastava2021diverse}. 

In video-to-video synthesis, motion information is provided in an explicit form, such as a set of sparse motion trajectories \cite{Hao_2018_CVPR}, a sequence of human poses \cite{vid2vid, Chan_2019_ICCV}, or a sequence of discrete actions \cite{Gafni2020Vid2Game, Menapace_2021_CVPR}. The spatial information is provided with separate images or frames (with or without segmentation masks) \cite{Hao_2018_CVPR, nips2019vid2vid, Menapace_2021_CVPR} or structured data such as 3D face mesh \cite{Zhao_2018_ECCV}. 
Thanks to the rich input information, video-to-video synthesis can generate high-quality videos with controllable motions. However, the motion sequence is usually hard to obtain. 
%Without the need of predicting object motions, video-to-video synthesis focus on motion transfer from source motion sequence to a target video \cite{Hao_2018_CVPR, vid2vid, nips2019vid2vid, Chan_2019_ICCV, Gafni2020Vid2Game, Zhou_2019_ICCV}. For example, Hao \textit{et al.} \cite{Hao_2018_CVPR} generate videos conditioned on a static image and a set of sparse motion trajectories. Wang \textit{et al.} \cite{vid2vid} learn a conditional generative adversarial model and successfully generate photorealistic videos from segmentation masks, sketches, and poses. They further introduce a network weight generation module to store appearance patterns from few example images, thus extending to few-shot video-to-video synthesis \cite{nips2019vid2vid}. Zhao \textit{et al.} \cite{Zhao_2018_ECCV} devise a two-stage generation framework incorporating with 3D face mesh. Chan \textit{et al.} \cite{Chan_2019_ICCV} extract keypoints-based pose as intermediate representation to transfer the motion of dance from a source video to a novel target. Menapace \textit{et al.} \cite{Menapace_2021_CVPR} introduce the playable video generation task to generate video given the first frame and a sequence of discrete actions. Because of the given motion sequences, video-to-video synthesis can generate highly controllable video, despite those motion sequences are often hard to get.

\subsection{Image-to-Video Generation}
Image-to-video generation refers to the line of work which generates video from a single image and a random or very coarse motion clue. 

When the motion clue is not provided at all, videos are generated in a stochastic manner constrained by the spatial information provided by the input image \cite{iclr2018, Xiong_2018_CVPR, Li_2018_ECCV, Yang_2018_ECCV, Castrejon_2019_ICCV, dtvnet, Dorkenwald_2021_CVPR}. %For example, MD-GAN \cite{Xiong_2018_CVPR} adopts two-stage network to generate videos of realistic contents for each frame and refine the temporal transformation, respectively, without incorporating any random noise. 
% Li et al. \cite{Li_2018_ECCV} synthesizes a set of future frames in a variational probabilistic manner from one single still image. 
The models used to generate videos can be generative adversarial network (GAN) \cite{dtvnet} or variational autoencoder (VAE) \cite{Dorkenwald_2021_CVPR}.
%DTVNet \cite{dtvnet} encodes optical flow maps into a normalized motion vector, which is further integrated into single image to generate target video based on GAN. Dorkenwald \textit{et al.} \cite{Dorkenwald_2021_CVPR} apply variational autoencoder to encode video representation and allow to sample a random vector to reconstruct scene dynamics through a conditional invertible neural network. 
This kind of stochastic video generation can only handle short dynamic patterns in distribution. 

In order to produce more controllable video, coarse-grained motion clues, including predefined directions or action labels, can be provided \cite{Pan_2019_CVPR}. %Yang \textit{et al.} \cite{Yang_2018_ECCV} propose a two-stage pose guided method which predict pose sequence first supervised by ground-truth human poses and then synthesize coherent video frames with input image. Pan \textit{et al.} \cite{Pan_2019_CVPR} take one semantic label map as input to synthesize a sequence of photo-realistic video frames. 
Recently, Blattmann \textit{et al.} \cite{Blattmann_2021_CVPR} propose an interactive I2V synthesis model which allows users to specify the desired motion through the manual poking of a pixel. % With auxiliary conditions supplying the initial dynamic direction, I2V can generate more controllable videos than stochastic generation. 
I2V generation does not require users to provide detailed motion information, which reduces the burden for use, but at the same time it is unlikely to be used for generating videos with complex motion patterns.

\subsection{Text-to-Video Generation}
T2V task aims at generating videos just from text descriptions, which is a challenging task. There is relatively little research on this topic.

Mittal \textit{et al.} \cite{10.1145/3123266.3123309} first introduced this task and proposed a VAE-based framework, called Sync-DRAW, to encode simple captions and generate semantically consistent videos. A concurrent work \cite{Marwah_2017_ICCV} performs variable-length semantic video generation from captions. The model relies on VAE and recurrent neural networks (RNNs) to learn the long-term and short-term context of the video. Other works \textit{et al.} \cite{mm2017, Li_Min_Shen_Carlson_Carin_2018, deng2019irc, balaji2019conditional} also try to generate video from the caption and a latent noise vector. However, due to the ambiguity of text and its inefficacy in providing fine-grained appearance information, the generated videos are usually at low resolution or very blurry. Recently, a VQ-VAE-based generator named GODIVA \cite{wu2021godiva} was proposed to generate open-domain videos from text. However, the appearance of the generated video tends to be the most salient feature that has been seen in the training phrase. It is unlikely to generate videos for complex scene or unseen appearance.

%Different from those tasks, the newly introduced TI2V task aims to generate video from an image and a text description.
Compared to I2V, the proposed TI2V can generate controllable video through maneuverable dynamics. Compared to V2V, TI2V does not need complex auxiliary information. Compared to T2V, TI2V can generate more controllable video due to the specified appearance and fine-grained motion.

\begin{figure*}[htbp]
    \centering
    \includegraphics[width=0.92\textwidth]{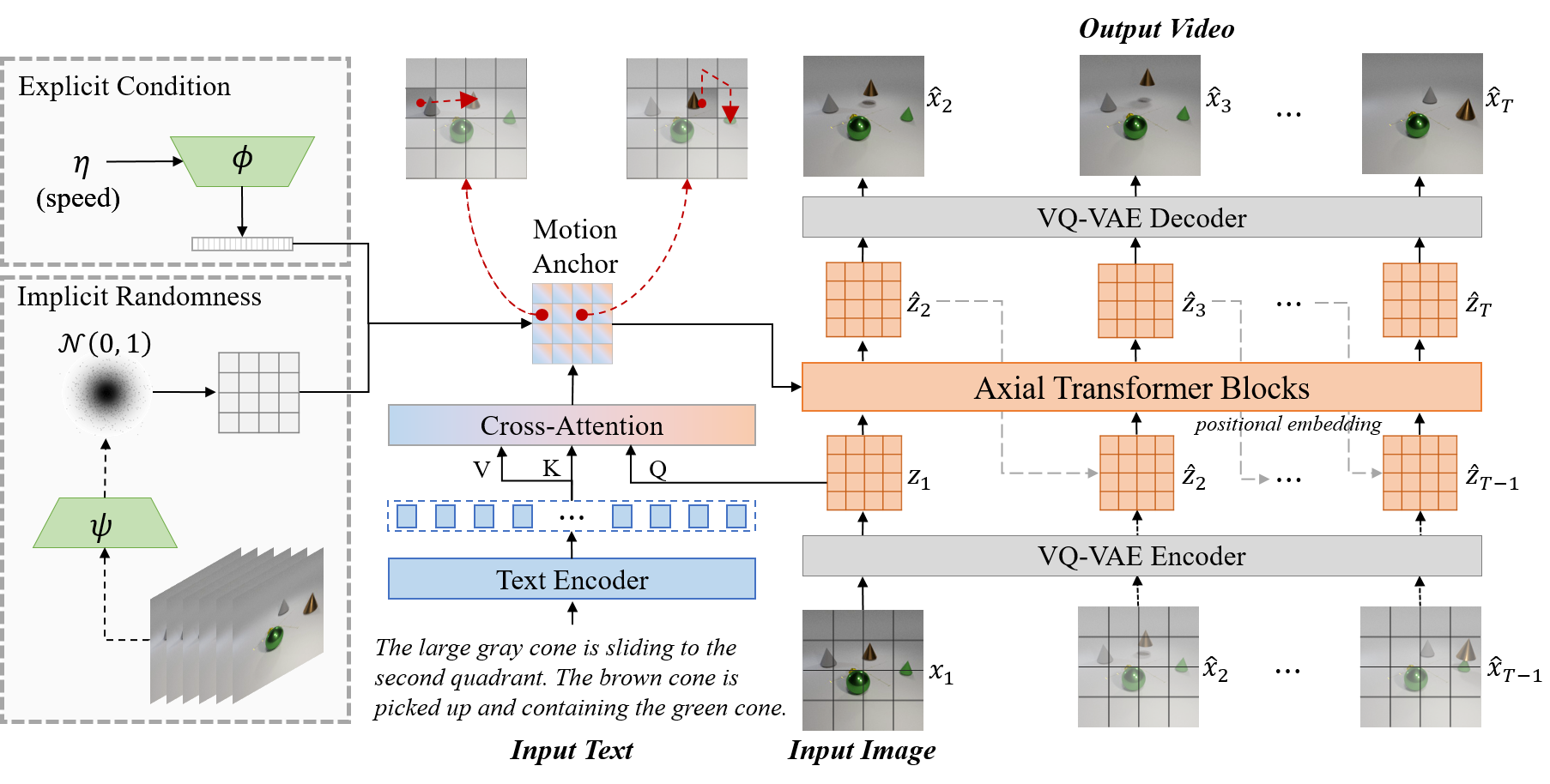}
    \caption{Illustration of the MAGE framework. The black dashed lines and grey dashed lines stand for operations that are only involved in training and inference processes, respectively. The black solid lines indicate the operations used in both processes. Each frame is represented by $4 \times 4$ tokens in the figure only for visualization purpose.}
    \label{fig:framework}
\end{figure*}

\section{MAGE for Text-Image-to-Video Task}
\label{sec:method}
\subsection{TI2V Problem Formulation}
TI2V task aims at generating a video from a static image and a text description. Formally, given a single static image $x_{1} \in \mathbb{R}^{h \times w \times C}$ and a text description $s=\left\{s_{1}, \cdots, s_{L}\right\}$ containing $L$ words, the goal of TI2V is to learn a mapping function that generates a sequence of frames $\hat{x}=\left\{\hat{x}_{2}, \cdots, \hat{x}_{T}\right\}$ with consistent appearance with $x_{1}$ and specified motion in $s$. 

We address the TI2V task through a supervised learning approach. During training, a reference video, denoted by $x=\left\{x_{2}, \cdots, x_{T}\right\}$ is provided for ($x_1$, $s$). The training objective is to make the conditional distribution of $\hat{x}$ given $x_{1}$ and $s$ approximate the conditional distribution of $x$. %given $x_{1}$ and $s$: $P\left(\hat{x} \mid x_{1}, s\right) \rightarrow P\left(x \mid x_{1}, s\right)$.

\subsection{MAGE Framework Overview}
MAGE adopts a VQ-VAE-based encoder-decoder architecture. VQ-VAE \cite{vqvae2017} is an effective tool to convert highly redundant visual data into a concise representation, which is a good choice for semantic-level manipulation, as we need to do in video generation. 

The entire framework is illustrated in Fig. \ref{fig:framework}. When an initial image $x_1$ and a text description $s$ is given, the image is passed to the VQ-VAE encoder and tokenized into a group (16x16) of latent codes $z_1$. The vector quantized image token, as well as text embeddings are then passed to the cross-attention module to obtain a spatially aligned motion representation, called motion anchor ($\widetilde{M}$).
% denoted by $z_1$ is then passed to the cross-attention module as the query (Q). The key (K) and the value (V) are the encoded text description. The output of the cross-attention module is called motion anchor ($\widetilde{M}$), which contains spatially aligned motion information for each token position. 
The explicit condition and implicit randomness, which will be detailed later, are also integrated into the MA. 

Then, the MA $\widetilde{M}$ is fused with $z_1$ by axial transformer blocks to produce $\hat{z}_2$, from which the VQ-VAE decoder can decode the next video frame $\hat{x}_2$. Once $\hat{z}_i$ ($i \ge 2)$ is obtained, it is sent back to the axial transformers to generate $\hat{z}_{i+1}$ together with $\widetilde{M}$ and all previous frames. Note that MA is a global variable which contains all the motion information needed to generate the entire sequence, so it only needs to be computed once. The generation process stops when a predefined sequence length is reached. 

The core of the MAGE framework is the motion anchor. All the related network models, including the cross-attention, the axial transformer blocks, the function $\phi$ that encodes explicit condition, and the function $\psi$ that encodes implicit randomness are trained together to achieve a unified objective. The VQ-VAE encoder and decoder, however, are peripheral modules which are trained beforehand. 

%The training process of MAGE consists of two stages: \uppercase\expandafter{\romannumeral1}) Training VQ-VAE for latent code learning. Parameters in the VQ-VAE Encoder and Decoder are frozen in the second stage. \uppercase\expandafter{\romannumeral2}) Training the other components for MA-based video generation. 
% We first quantize the RGB image $x$ to latent space token $z$ in stage \uppercase\expandafter{\romannumeral1} to reduce the computation cost. In stage \uppercase\expandafter{\romannumeral2}, the motion anchor (MA) is extracted and incorporates explicit condition and implicit randomness. Then the video generator, made up of axial transformer blocks, jointly models the MA and visual tokens to generate future frames recursively. We now describe these two stages in detail.

For completeness, we first introduce how we pre-train VQ-VAE. A VQ-VAE model consists of a latent codebook $C \in \mathbb{R}^{K \times D}$, an encoder $E$ and decoder $D$ with respective down-sampling and up-sampling ratio $n$.
%which contains $K$ embedding vectors $b_{i} \in \mathbb{R}^{D}, i \in 1,2, \ldots, K$.
The input image $x \in \mathbb{R}^{H \times W \times C}$ is encoded into latent vector $e_{x}=E(x) \in \mathbb{R}^{h \times w \times D}$ first, where $h=H / n, w=W / n$. Then $e_{x}$ is discretized by a nearest neighbour look-up in $C$ to get both quantized index $z \in \mathbb{R}^{h \times w}$ and quantized vector $\tilde{e}_{x} \in \mathbb{R}^{h \times w \times D}$. 
The decoder, with a reversed structure of the encoder, then reconstructs image $\hat{x}$ from $\tilde{e}_{x}$.
% After VQ, the amount of video data $X \in \mathbb{R}^{T \times H \times W \times C}$ is greatly reduced to $Z \in \mathbb{R}^{T \times h \times w}$. 
%Besides, it also releases the burden of pixel-level reconstruction for video generation model.

VQ-VAE is trained with an image-level reconstruction task. The training objective consists of the reconstruction loss, codebook loss, and commitment loss. It is written as:
\begin{equation}
    \mathcal{L}=\log P\left(\hat{x} \mid \tilde{e}_{x}\right)+\left\|\operatorname{sg}\left(e_{x}\right)-\tilde{e}_{x}\right\|_{2}^{2}+\beta\left\|e_{x}-sg\left(\tilde{e}_{x}\right)\right\|_{2}^{2}, 
\end{equation}
where $\operatorname{sg}$ stands for the stop-gradient operator and $\beta$ is the weighting factor. After this pre-training stage, parameters in $E$ and $D$ are frozen. 
%The first reconstruction loss is to encourage the latent representations to possess reconstruction ability. The second codebook loss aims to decrease the distance between latent embedding and quantized embedding. The last commitment loss constrains the encoder to be more stable and informative.

\subsection{MA-Based Video Generation}
Motion anchor is the core design in MAGE to achieve controllable and diverse video generation for the TI2V task. It aligns text with image, and allows for the injection of explicit condition and implicit randomness. 

\subsubsection{Image-Text Alignment}
In MAGE, we employ a cross-attention module to achieve the alignment between the image content and the motion clues given in text. We employ a learnable text encoder to compute the text embedding $e_{s} \in \mathbb{R}^{L \times d}$ from the input text $s$, where $d$ is the hidden size. $e_s$ is used as the key and the value of the cross-attention module. The image embedding $z_1$ is converted to the same latent space as $e_s$ by a learnable embedding matrix, and the converted embedding, denoted by $e_{z_{1}} \in \mathbb{R}^{h \times w \times d}$, is used as the query. 

The cross-attention operation locates the responsive words for each visual token and aggregates the implied motion information. Then, the motion information and the corresponding visual information are fused by a feed forward and normalization layer to generate the motion anchor $M \in \mathbb{R}^{h \times w \times d}$. This process can be described by: 

\begin{equation}
\begin{aligned}
&Q=e_{z_{1, i, j}} W^{q}, K=e_{s} W^{k}, V=e_{s} W^{v},\\
&A_{i, j}=\text { MultiHead }(Q, K, V), \\
&M_{i, j}=F F N\left(e_{z_{1, i, j}}, A_{i, j}\right).
\end{aligned}
\end{equation}
where $e_{z_{1, i, j}}$ stands for the visual embedding at position $\left(i, j\right)$ in the first frame. $M_{i, j}$ stores the appearance and motion information at position $\left(i, j\right)$. %Besides, the spatially aligned structure of MA also enables the injection of external prior both on local and global.

\subsubsection{Explicit Condition and Implicit Randomness}
We allow MA to encode some quantifiable conditions that are not expressed in the text. In this work, we demonstrate how a typical condition known as motion speed is incorporated into the generation process and reflected in the generated video. 
% To enhance the representation of MA, additional explicit conditions are allowed to be modeled jointly. Considering that one drawback of existing video generation model is that it can only generate short and length-limited video clips because of large data quantity and limited model capacity, the completion of motion is constrained by video length and cannot be controlled. Thus, we propose a compromise by selecting speed as explicit condition $\eta$ to improve both controllability and diversity. To encode variable $\eta$, 
A simple linear layer $\phi$ is applied to encode speed $\eta$ into an embedding vector $c \in \mathbb{R}^{d}$. It is written as:
\begin{equation}
    c=\phi(\eta).
\end{equation}

Besides, text description can be ambiguous. The ``correct" video that matches an input image-text pair may not be unique. Thus, the model is required to accommodate existing randomness $r$ in data distribution and randomly generate diverse videos which are semantically consistent with the text. %The implicit randomness $r$ is modeled under two constrains: \expandafter{\romannumeral1}) $r$ should adequately characterize the motion randomness existing in data, \expandafter{\romannumeral2}) $r$ should be compact and contain little irrelevant information. 
We propose to use a variational information bottleneck \cite{vib2017} $\psi$ for implicit randomness modeling. $\psi$ consists of several 3D convolutional blocks and a reparameterization layer. % followed by a 2D convolutional layer. 
During training, we encode the video randomness into a random variable that conforms to the standard normal distribution. During inference, a random variable is directly sampled from the distribution and merged into the MA, %Then the randomness $r$ is sampled from the predicted distribution, which is assumed to accord with the standard normal prior,
\begin{equation}
    r \sim q_{\psi}\left(r \mid e_{z_{1} \sim T}\right).
\end{equation}

To inject randomness $r$ into MA, an adaptive instance normalization (AdaIN) layer \cite{huang2017arbitrary} is applied. Since speed affects each movement equivalently, $c$ is directly injected into $M$ through  and a channel-wise additive to change the global motion information. It is formulated as
\begin{equation}
    \widetilde{M}=A d a I N(M, r)+c.
\end{equation}
% With the AdaIN layer, MA is perturbed by $r$ on mean and variance. Given the same image and text input, our model can generate diverse videos by sampling different random noise.  
% The updated spatially aligned MA can be easily aligned with frame tokens in the video generator $G$ further to generate coherent videos.

\subsubsection{Appearance-Motion Fusion}
After obtaining the motion anchor $\widetilde{M} \in \mathbb{R}^{h \times w \times d}$, the video generator $G$ jointly models MA and visual token embeddings. To reduce computation, we adopt $N$ axial transformer blocks which consist of three-dimensional axial attention \cite{ho2019axial, wu2021godiva} on temporal-wise, row-wise, and column-wise, respectively. As such, the attention complexity is reduced from $\mathcal{O}\left((T h w)^{2}\right)$ to $\mathcal{O}\left(Thw\left(T+h+w\right)\right)$. The generation can be formulated as
\begin{equation}
    \hat{z}_{i}=G\left(PE\left(\left[\widetilde{M} \cdot z_{<i}\right]\right)\right),
\end{equation}
where $\left[ \cdot \right]$ stands for the concatenation operation and $PE$ represents positional embedding. Noted that row-wise and column-wise attention have full receptive field on respective axis. But for temporal-wise axial attention, we apply a causal mask to ensure that a visual token can only receive the information from previous frames. After stacking several axial transformer blocks, each visual token has full receptive field on current and previous frames for spatial and temporal information. A token in each position can not only get the complete motion information from the MA, but also ``track" the motion in previous frames. Recurrently generating frames with spatially aligned MA ensures coherent and consistent video output.

The training objective for video generator consists of a cross-entropy loss for visual token prediction, and two constraints for explicit condition $c$ and implicit randomness $q_{\psi}(r\mid X)$. It is formulated as
\begin{equation}
\begin{aligned}
    \mathcal{L}=& -\frac{1}{T} \sum_{i=2}^{T} z_{i} \log \left(P\left(\hat{z}_{i} \mid z_{<i}, s, c, r\right)\right)\\
    & +\alpha\|c\|_{2}^{2}+\beta K L\left(q_{\psi}(r\mid X) \| p(r)\right),
\end{aligned}
\end{equation}
where $\alpha$ and $\beta$ are hyper-parameters to trade-off the two constraints, and $p(r)$ stands for the standard normal prior.

\section{Experiments}

\subsection{Datasets}
Traditional T2V methods are often evaluated on Single Moving MNIST and Double Moving MNIST \cite{10.1145/3123266.3123309} that contain one and two moving digits, respectively. Four motion patterns are included: right then left, left then right, up then down, down then up. Once the border is reached, the digit will rebound. 
%However, an important characteristic of TI2V task is the control of text over the motion process. The motion patterns in Single/Double Moving MNIST only need an initial direction and then move endlessly with constant bounce, simplifying the video generation into I2V task given one image and a predefined direction. To improve the controllability of text, 
We propose a Modified Double Moving MNIST dataset with more movements and a distracting digit. We also propose two versions of synthetic CATER-GEN datasets in a 3D environment with lighting and shadows built upon CATER \cite{Girdhar2020CATER}. These datasets are briefly introduced as follows:
\begin{figure*}[htbp]
    \centering
    \includegraphics[width=0.95\textwidth]{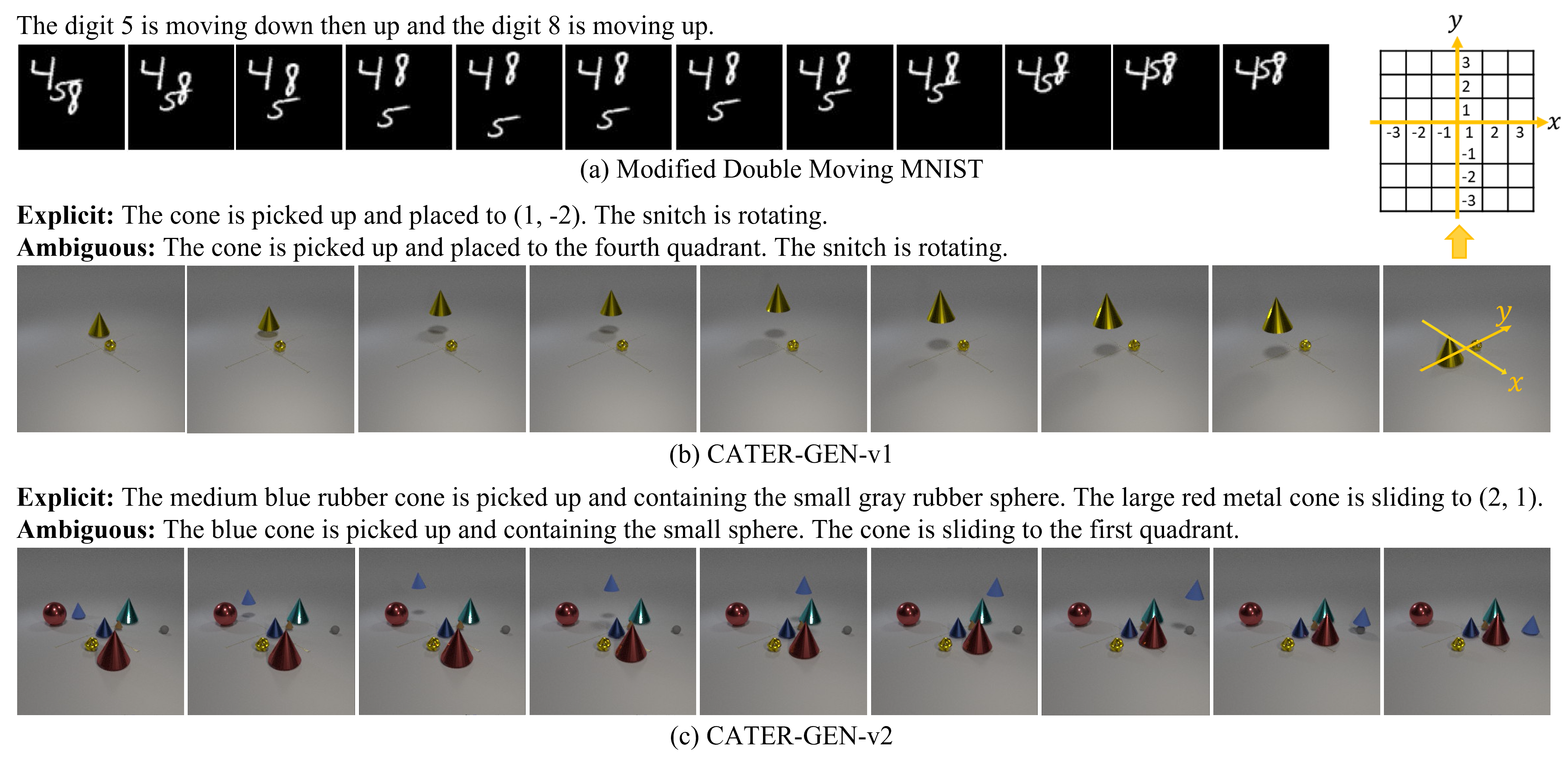}
    \caption{Samples from Modified Double Moving MNIST, CATER-GEN-v1 and CATER-GEN-v2, respectively. For CATER-GENs, we split the 2D table plane into a $6 \times 6$ portion with fixed axes as shown on the top right. Thus, the position of object can be described with coordinate or quadrant. Note that the camera position is static.}
    \label{fig:dataset}
\end{figure*}
\begin{itemize} 
    \item \textbf{Modified Double Moving MNIST}: In order to control the moving process, we keep two options for the movement along each direction. One is to stop at the edge and the other is to bounce once. Once the action finishes, the digit will stay still. Thus, we get 8 motion patterns for all four directions. Besides, we randomly insert one static distracting digit other than the moving ones as background.
    \item \textbf{CATER-GEN-v1}: CATER-GEN-v1 is a simpler version which is built with two objects (cone and snitch) and a large "table" plane inherited from CATER. There exist four atomic actions: ``rotate", ``contain", ``pick-place" and ``slide". Each video randomly contains one or two actions. When generating descriptions, we design a predefined sentence template to fill the subject, action, and optional object. The final position is also provided for actions ``pick-place" and ``slide". By specifying the final position with an accurate coordinate or a quadrant area, explicit descriptions and ambiguous descriptions are provided for deterministic and stochastic video generation, respectively.
    \item \textbf{CATER-GEN-v2}: CATER-GEN-v2 is a much more complex dataset which contains $3\sim8$ objects in each video. Each object has 4 attributes that are randomly chosen from five shapes, three sizes, nine colors, and two materials. The atomic actions are the same as in CATER-GEN-v1. To create ambiguity in text descriptions, we not only replace the final coordinate, but also randomly discard the attributes for each object, thus the object may not be unique due to the uncertainty in the referring expression.
\end{itemize}

The resolution of the generated video is $64\times64$ for the new and two existing MNIST-based datasets, and $256\times256$ for two CATER-based datasets. Both Single Moving MNIST and Double Moving MNIST contain 10k pairs for training and 2k for testing following \cite{deng2019irc}. For CATER-GEN-v1, we generate 3.5k pairs for training and 1.5k pairs for testing. For the more complicated datasets Modified-MNIST and CATER-GEN-v2, we generate 24k pairs for training and 6k pairs for testing. Samples from generated datasets are shown in Fig. \ref{fig:dataset}.

\subsection{Implementation Details}
Both VQ-VAE and video generator in our experiments are trained from scratch. For VQ-VAE, we use the similar encoder and decoder structure as \cite{ramesh2021zeroshot} with codebook size $512 \times 256$. The input size is $H=64, W=64, C=1$ for MNIST-based datasets, and $H=128, W=128, C=3$ for CATER-based datasets. After VQ, images are compressed to $16 \times 16$ visual tokens. The text description is encoded by a two-layer transformer. We let $T=10, d=512$ and stack two axial transformer blocks to generate the video. We use a batch size of 32 and a learning rate of 5e-5. The speed $\eta$ is normalized to $(0,1)$, which is mapped to a predefined frame sampling interval.

In the training stage, the speed $\eta$ is randomly sampled from $(0,1)$ and we use the corresponding sampling interval to obtain the reference video. To improve training efficiency, we input $s$ and $\left\{x_{1}, \cdots, x_{T-1}\right\}$ to predict future frame tokens $\left\{\hat{z}_{2}, \cdots, \hat{z}_{T}\right\}$ in parallel. For inference, only the first image and a text description are given. We sample the implicit randomness from a normal distribution and generate the video sequence in an auto-regressive way.

\subsection{Deterministic Video Generation}
To evaluate whether MAGE can achieve the first goal of TI2V task, known as controllability, we discard implicit randomness module and use the explicit descriptions in the dataset. Thus, the whole model is deterministic, and the ``correct" video is unique. In this section, we first show the qualitative performance of generated videos (please refer to supplementary Sec.\ref{appendix:sec2} for quantitative results and ablation studies), then the controllability of explicit condition is evaluated. More interestingly, we try to input the same image but different descriptions (speed kept unchanged), to show how to manipulate the objects through different text.

\begin{figure*}[t]
    \centering
    \includegraphics[width=0.96\textwidth]{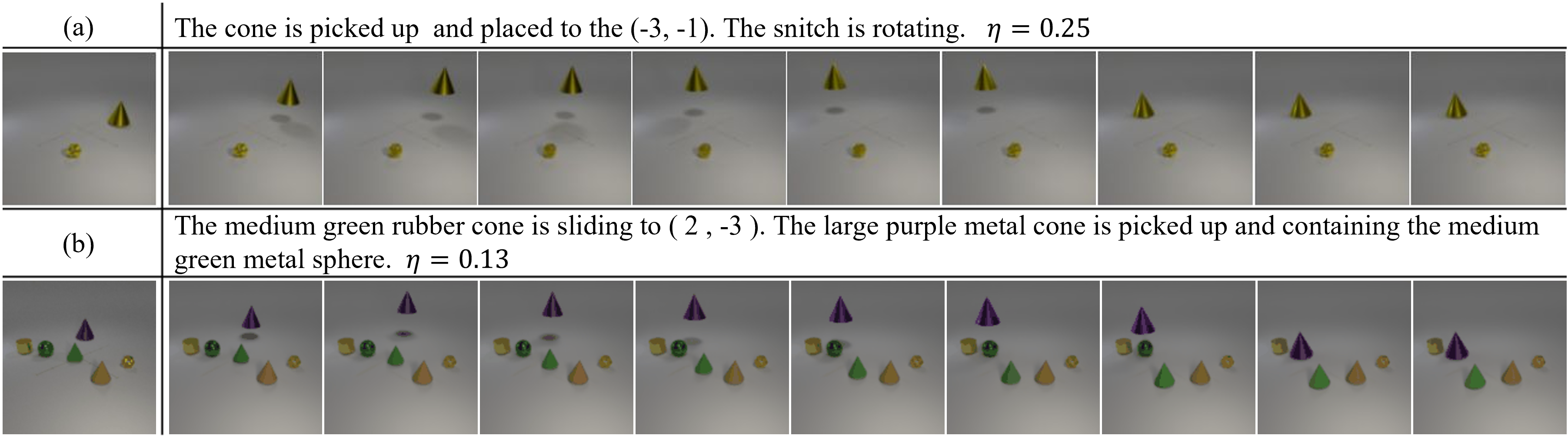}
    \caption{Generated Samples from CATER-GEN-v1 and CATER-GEN-v2, respectively. The images of the left column are given first images, and the first row of each sample is the given text and speed.}
    \label{fig:results1.2}
\end{figure*}

\begin{figure}[t]
    \centering
    \includegraphics[width=0.478\textwidth]{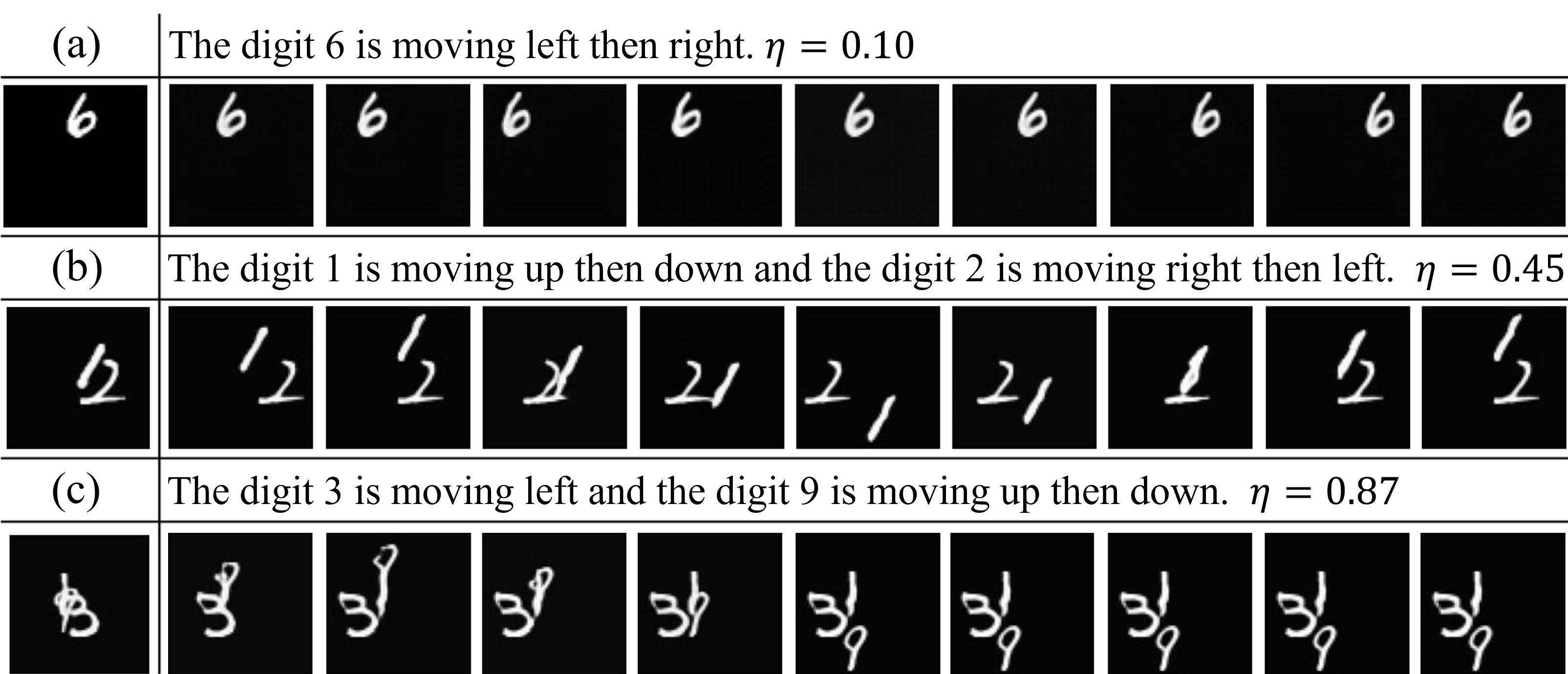}
    \caption{Generated Samples from Single / Double / Modified Double Moving MNIST, respectively, from top to bottom. The left column shows given images, and the first row of each sample is the given text and speed.}
    \label{fig:results1.1}
\end{figure}

\subsubsection{Qualitative Evaluation}
We first show the qualitative results on three MNIST-based datasets and two CATER-based datasets with explicit descriptions in Fig.\ref{fig:results1.1} and Fig.\ref{fig:results1.2}, respectively. The generated videos have high visual quality, and the motion is highly coherent with the one specified in the text. For three MNIST-based datasets, the digits are moving smoothly and the shapes are well maintained. Especially for the hard example from Modified Double Moving MNIST (Fig.\ref{fig:results1.1}(c)), the given image contains three digits with a large overlap. MAGE is able to recognize ``3" and ``9" specified in the text and disassemble them, while keeping the unmentioned ``1" still. 

In CATER-GEN-v1, although there are only two objects, it is still a challenge to not only generate right motion, but also simulate variations of surface lighting and shadows of objects because of three illuminants out of scene. It can be observed in Fig.\ref{fig:results1.2}(a) that the cone is placed to the right coordinate. Both the changing of surface light and shadow are generated quite well. For the rotating of snitch, it is difficult for both VQ-VAE and the generation model to reconstruct such small object with complex appearance. We can still observe the variation on the surface which suggests that there is an action ``rotate" happening. 

For the sample from CATER-GEN-v2 as shown in Fig.\ref{fig:results1.2}(b), although there are six different objects, the model successfully locates the ``medium green rubber cone", ``large purple metal cone", and ``medium green metal sphere". It can also handle the occlusion relation for action ``contain". Meanwhile, other unmentioned objects are kept stable and still.

\subsubsection{Explicit Condition Evaluation}
\begin{figure}[t]
    \centering
    \includegraphics[width=0.37\textwidth]{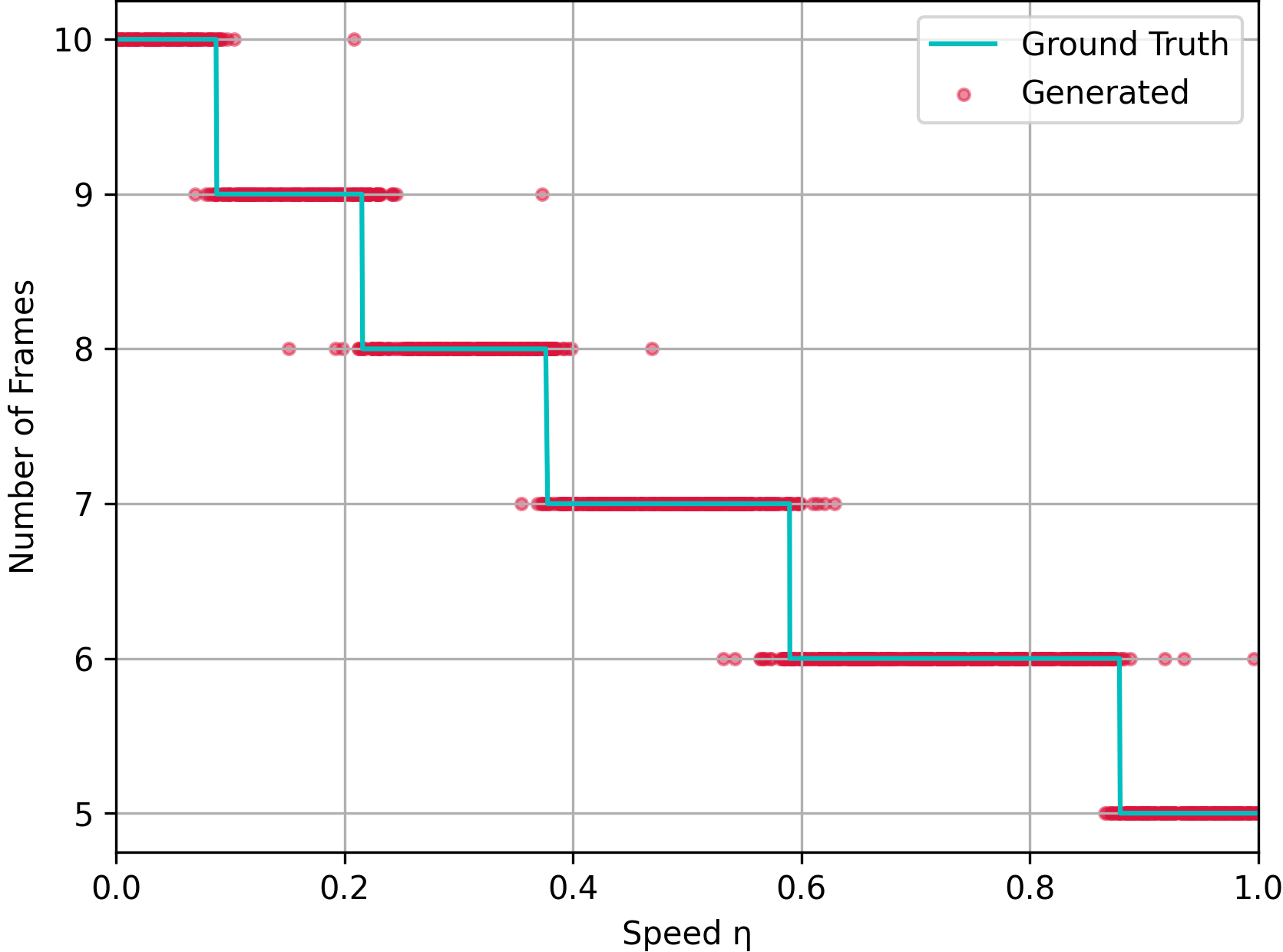}
    \caption{Statistics of the number of frames when the motion finishes from CATER-GEN-v1 test set (1500 samples), where the X-axis stands for the input speed $\eta$ and Y-axis stands for the number of frames. The blue line and red scatter represent ground truth and generated video, respectively.}
    \label{fig:results2}
\end{figure}

\begin{figure*}[t]
    \centering
    \includegraphics[width=1.0\textwidth]{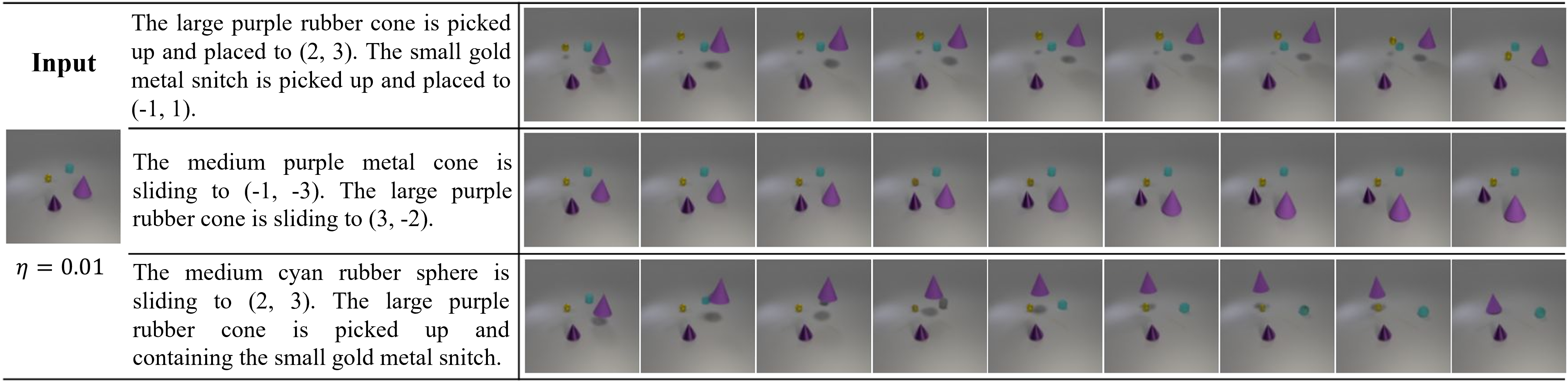}
    \caption{Generated Samples from CATER-GEN-v2 for composability evaluation. The left column shows the input with the same image and speed but different texts.}
    \label{fig:results3.2}
\end{figure*}

To evaluate the effectiveness in handling the explicit condition, we count the number of frames when the motion finishes under different speed settings. Fig.\ref{fig:results2} shows the result on CATER-GEN-v1 with predefined sampling interval $(3,6)$. It can be found that the speed of generated video is consistent with ground truth. The borderline speed may cause confusion due to the sampling accuracy, but the error is within one frame. It proves that the MA is able to integrate the information specified in the explicit condition and pass it to visual tokens. We believe that this structure also works for other quantifiable conditions. % showing the potential of drawing other conditions into video generation model by simply changing the input $\eta$.

\begin{figure}[t]
    \centering
    \includegraphics[width=0.46\textwidth]{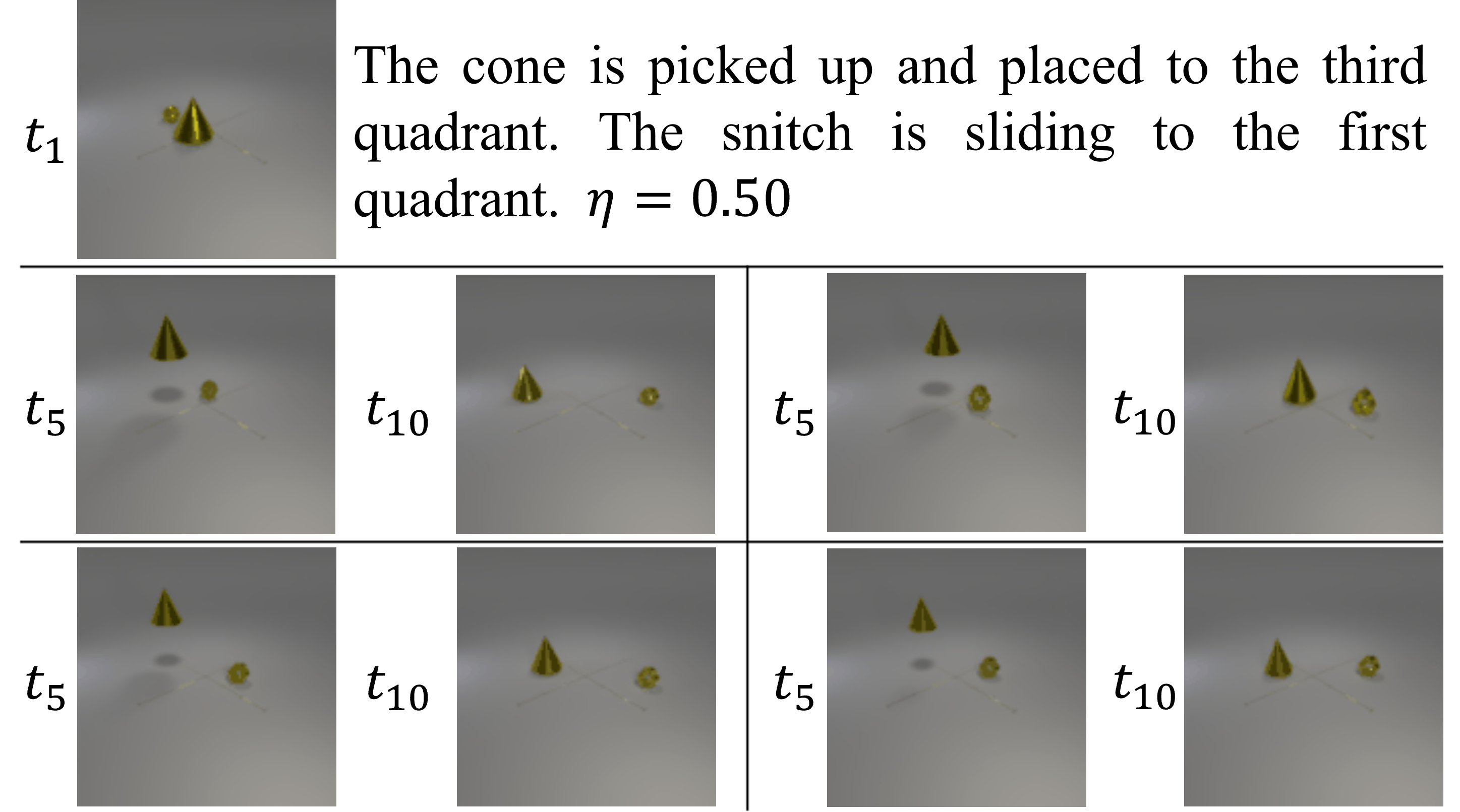}
    \caption{Generated Samples of stochastic video generation from CATER-GEN-v1. Given the same image, description and speed as input in the first row, we generate four videos and show the 5$th$ and 10$th$ frames.}
    \label{fig:results4.1}
\end{figure}

\subsubsection{Composability Evaluation}
In order to show the controllability of text, we try to use different descriptions to manipulate the same image. 
We show the results of CATER-GEN-v2 in Fig.\ref{fig:results3.2} trying to manipulate different objects moving to different positions. The results indicate that our model can recognize different objects as well as their attributes like ``medium purple metal cone" and ``large purple rubber cone". In our experiments, we observe in some rare cases that the color of objects may change in the generated videos (the cyan sphere in the last row of Fig.\ref{fig:results3.2}). This is actually caused by the reconstruction quality of VQ-VAE. We expect that this error will disappear when a higher-capacity VQ-VAE is trained. %Since the reconstruction accuracy of VQ-VAE directly affect the performance of generated videos, resulting that the generation model may switch the color of objects .

\subsection{Diverse Video Generation}
The uncertainty in description is a great challenge for TI2V task. In this experiment, we try to generate video with ambiguous descriptions to show that whether the model can generate diverse videos under the premise of semantically consistency with text. We conduct experiments on CATER-GEN-v1 and CATER-GEN-v2 with ambiguous descriptions and add the implicit randomness module. In the inference phrase, multiple videos are generated for the same set of image, text, and speed as input. For CATER-GEN-v1 dataset, the randomness implied in the description is the final position for action ``pick-place" and ``slide", as only quadrant is provided instead of exact coordinate. The results shown in Fig.\ref{fig:results4.1} indicate that our generation model can generate diverse videos in which objects are placed to the correct quadrant but random positions. 

\begin{figure}[t]
    \centering
    \includegraphics[width=0.46\textwidth]{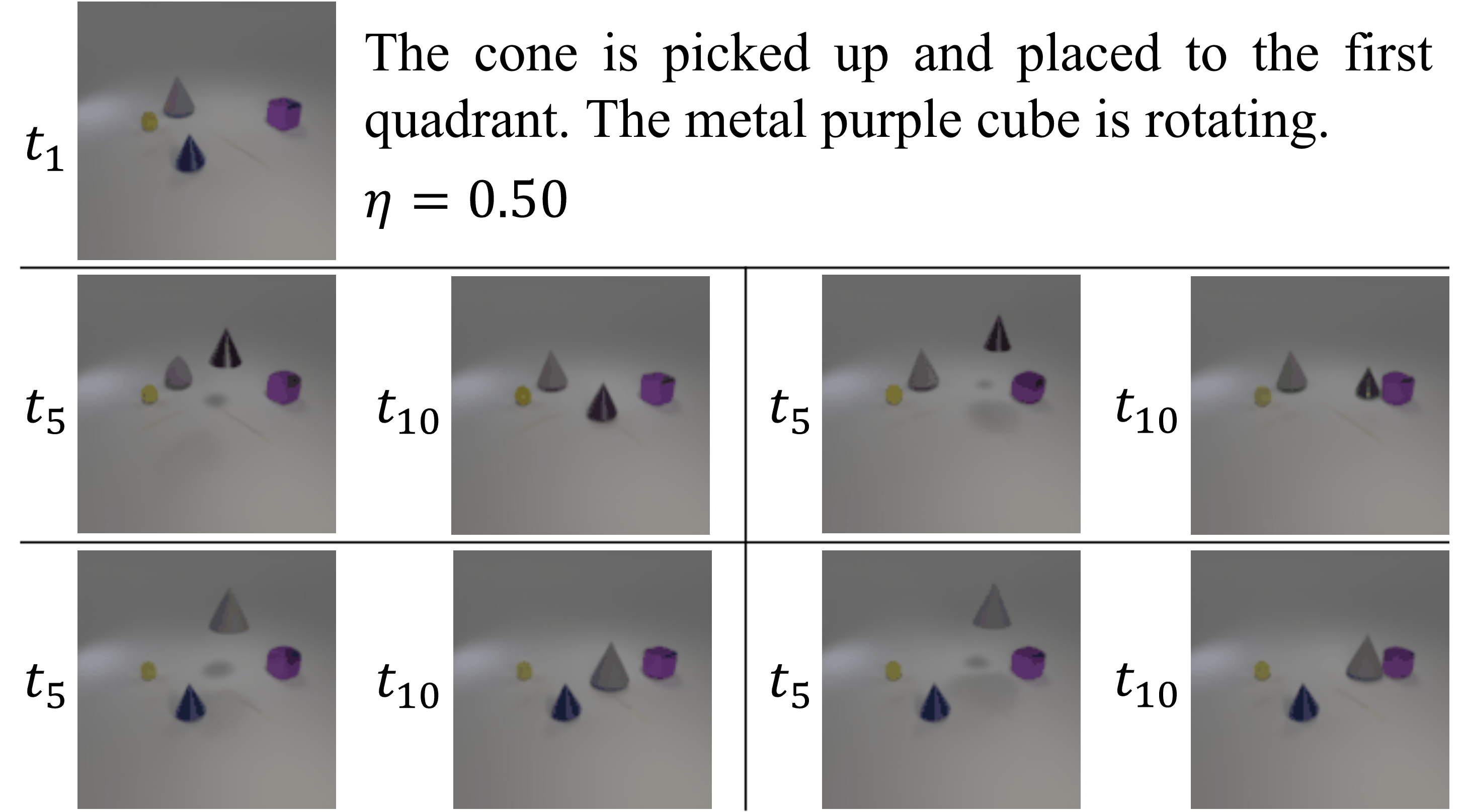}
    \caption{Generated Samples of diverse video generation from CATER-GEN-v2.}
    \label{fig:results4.2}
\end{figure}

For CATER-GEN-v2 dataset, the implicit randomness exists in both position and referring expression. We show an example in Fig.\ref{fig:results4.2} that the given image contains two cones that are not distinguished in the description. The generated results show that both the gray cone and the blue cone have chances to be picked. In addition, the placement positions are all in the first quadrant but can be different coordinates. 

The stochastic video generation results indicate that the uncertainty existed in data distribution can be automatically extracted and integrated into motion anchor, while generating reasonable and diverse videos. It is encouraging to see the success of MAGE on these two datasets with controlled randomness. It suggests that MAGE may also work well for other challenging datasets if the randomness can be efficiently modelled. Please refer to the supplementary Sec.\ref{appendix:sec3} for more visualizations.

\section{Conclusion}
In this paper, we have introduced a novel Text-Image-to-Video generation task aiming to generate video from a static image and a text description. We have proposed MAGE which based on motion anchor enabling alignment between appearance and motion representation to generate controllable and diverse videos. To evaluate the feasibility of TI2V task and our model, we have also introduced the Modified Double Moving MNIST and CATER-GEN datasets. Experiments have shown that our model can not only generate coherent and consistent videos, but also successfully model both explicit condition and implicit randomness. However, it is still a great challenge to generate realistic and open-domain videos because of the high randomness and diversity in video-text pairs. We believe TI2V is a challenging and valuable research direction. In the future, we will extend MAGE to more challenging realistic video data.

% %------------------------------------------------------------------------
% \section{Final copy}

% You must include your signed IEEE copyright release form when you submit your finished paper.
% We MUST have this form before your paper can be published in the proceedings.

% Please direct any questions to the production editor in charge of these proceedings at the IEEE Computer Society Press:
% \url{https://www.computer.org/about/contact}.

%%%%%%%%% REFERENCES
{\small

\bibliographystyle{ieee_fullname}
%\bibliography{egbib}
}

\clearpage
% \newpage
\appendix
\begin{Large}
\noindent \textbf{Supplementary Material}
\end{Large}
\section{Dataset Generation}
\label{appendix:sec1}
\subsection{Modified Double Moving MNIST}
This dataset contains 8 motion patterns combined by 4 directions and 2 modes (no bounce or bounce once). In order to improve the controllability of text and make the dataset more complex, we randomly add one static digit as background. To avoid the ambiguity of description, two digits of the same number are not allowed to exist in one image. 

Following the Single/Double Moving MNIST, the combinations of digit and motion pattern in training and testing set are mutually exclusive. That means, for example, digit 9 only moves horizontally in training set, but vertically in testing test. 

\subsubsection{CATER-GEN-v1}
CATER-GEN-v1 is a smaller and simpler version of CATER-GENs to facilitate the observation of not only actions of generated video, but also the variation of surface lighting, the shadow, and the background. There are three illuminations out of scene including key light, fill light, and back light. Once the object moves, the surface light and shallow will also change, bringing a challenge for reasonable and accurate video generation. This dataset only contains two objects: cone and snitch (like three intertwined tori) in metallic gold color. The initial position of objects on the table plane is randomly selected from a $6 \times 6$ portion. We inherit four actions in CATER: ``rotate", ``pick-place", ``slide", and ``contain". Each video randomly contains one or two actions that happen at the same time. The ``rotate" action is afforded by snitch. For ``pick-place" and ``slide" actions, the target position is also randomly selected. We define the descriptions according to shapes, actions, and coordinates (``the cone is picked up and containing the snitch", ``the snitch is sliding to (-1, 3)"). We also provide the version of ambiguous descriptions by replacing the coordinate with quadrant for diverse video generation (``the snitch is sliding to the second quadrant").

\subsubsection{CATER-GEN-v2}
Based on the pipeline of CATER dataset, we inherit the objects and actions. Specifically, objects include five shapes (cube, sphere, cylinder, cone, snitch), in three sizes (small, medium, large), two materials (metal, rubber) and nine colors (red, blue, green, yellow, gray, brown, purple, cyan, and the gold only for snitch). The snitch is a special object with fix size, material, and color. The ``rotate" action is afforded by cubes, cylinders and the snitch, while the ``contain" action is only afforded by the cones. In CATER-GEN-v2, each video randomly contains one or two actions that both start at the first frame. We fix the camera position to ensure the consistency of coordinate system. To generate explicit descriptions, we provide all properties, action, and coordinate for each moving object like ``the medium blue rubber cone is picked up and placed to (1, -3)". To generate ambiguous descriptions, we randomly discard attributes (size, material, color) for each object thus brings into the uncertainty of referring expression. Like CATER-GEN-v1, we also replace the coordinate with quadrant to produce the uncertainty of movements (``the rubber cone is picked up and placed to the fourth quadrant").

\section{Quantitative Results}
\label{appendix:sec2}

\begin{table*}[htbp]
\newcommand{\tabincell}[2]{\begin{tabular}{@{}#1@{}}#2\end{tabular}}
    \centering
    \begin{tabular}{cc|ccc|cc}
        \toprule
        Mode & Datasets & FID $\downarrow$ & LPIPS $\downarrow$ & FVD $\downarrow$ & DIV VGG $\uparrow$ & DIV I3D $\uparrow$\\
		\midrule
        \multirow{2}[1]{*}{\tabincell{c}{Deterministic\\(explicit text)}} & CATER-GEN-v1 & 62.66 & 0.20 & 31.70 & 0 & 0\\
        % \cmidrule{2-5}
        & CATER-GEN-v2 & 39.56 & 0.20 & 57.55 & 0 & 0\\
        \midrule
         \multirow{2}[1]{*}{\tabincell{c}{Stochastic\\(ambiguous text)}} & CATER-GEN-v1 & 62.89 & 0.22 & 45.49 & 0.15 & 0.45\\
        % \cmidrule{2-5}
        & CATER-GEN-v2 & 39.38 & 0.26 & 69.44 & 0.37 & 2.06\\
         \bottomrule
    \end{tabular}
    \caption{Qualitative results on CATER-based datasets under deterministic and stochastic video generation, respectively.}
    \label{tab:2}
\end{table*}

Since video generation is a challenging and relatively new task, there are few effective metrics to evaluate generated videos currently. To quantitatively evaluate MAGE, we apply conventional pixel-based similarity metrics SSIM and PSNR for deterministic video generation. We also report several perceptual similarity metrics including image-level Fr\'echet Inception Distance (FID)\cite{heusel2017gans} and Learned Perceptual Image Patch Similarity (LPIPS)\cite{dosovitskiy2016generating}, as well as video-level Fr\'echet-Video-Distance (FVD)\cite{unterthiner2018towards}. To evaluate the diversity of generated videos given ambiguous text, following previous work \cite{Dorkenwald_2021_CVPR}, we measure the average mutual distance of generated video sequences in the feature space of both VGG-16 \cite{vgg16} and I3D \cite{szegedy2016rethinking} network. The VGG and I3D backbones used in similarity and diversity metrics are pre-trained on ImageNet \cite{russakovsky2015imagenet} and Kinetics \cite{kay2017kinetics}, respectively.

\begin{table}[htbp]
    \centering
    \begin{tabular}{p{105pt}<{\centering}|p{20pt}p{30pt}p{32pt}}
        \toprule
        \multirow{2}[1]{*}{Datasets} & \multirow{2}[1]{*}{SSIM $\uparrow$} & \multicolumn{2}{c}{PSNR $\uparrow$} \\
         & & VQ-VAE & MAGE \\
		\midrule
         Single Moving MNIST & 0.97 & 43.12 & 33.89 \\
         Double Moving MNIST & 0.87 & 38.80 & 24.66 \\
         Modified Moving MNIST & 0.85 & 37.63 & 23.24\\
        \midrule
         CATER-GEN-v1 & 0.97 & 47.01 & 35.03 \\
         CATER-GEN-v2 & 0.95 & 40.21 & 32.74 \\
         \bottomrule
    \end{tabular}
    \caption{Qualitative results under deterministic video generation.}
    \label{tab:1}
\end{table}
Due to the novel setting of TI2V task, there exists a restrictive relation between high similarity and large diversity. Generated videos are required to be more diverse and semantically consistent with text at the same time. It is hard to fairly compare with other methods for I2V or T2V task, as methods for I2V task fail to generate controllable video with complicated motion in text. And methods for T2V task tend to generate correlated visual features that have been seen in the training stage, making it difficult to generate video from a novel image and model the uncertainty in text. 

Tab.\ref{tab:1} shows the PSNR and SSIM results on all datasets under deterministic video generation that only involves explicit text descriptions. In the testing stage, the speed $\eta$ is randomly sampled from $(0,1)$ for each sample. As the video generation performance is based on the reconstruction accuracy of VQ-VAE, both PSNR results of reconstructed video form VQ-VAE only and generated videos from MAGE are reported against ground truth videos. It can be found that our generated videos achieve high similarity with ground truth videos. When the dataset becomes harder, the video generation performance is considerably affected by VQ-VAE and declines. 

To further evaluate the ability to handle ambiguous text, we compare the perceptual similarity and diversity after applying implicit randomness module. When calculating diversity metric, we fix the speed input and generate 5 video sequences for each sample. The results are shown in Tab.\ref{tab:2}. Given explicit text, the generated video is unique and shows high similarity with reference video. After replacing explicit text with ambiguous text, the ground-truth video is no longer unique in this situation. The similarity between generated video and reference video decreases within an acceptable range. At the same time, the model is able to generate diverse videos. The results of CATER-GEN-v2 show much higher diversity than CATER-GEN-v1, which is also consistent with different degrees of uncertainty in text.

In addition, to validate the effectiveness of the cross-attention of MA and axial transformers, we conduct ablation study to compare with concatenation operation and vanilla transformer, respectively, in Tab. \ref{tab:ablation}. First, the 1st and the 2nd rows show that cross-attention in MA has better performance than concatenation with 1.4 decrease on FID while keeping similar LPIPS and FVD. Then, comparing the 2nd and the 3rd rows shows that axial transformers incur some performance loss but it remarkably reduces the computational complexity of vanilla transformers by $46\%$, which is consistent with our expectation.

% \vspace{-0.6em}
\begin{table}[htbp]
\footnotesize
\newcommand{\tabincell}[2]{\begin{tabular}{@{}#1@{}}#2\end{tabular}}
    \centering
    \begin{tabular}{p{13pt}p{13pt}|p{13pt}p{14pt}|p{10pt}p{13pt}p{13pt}|p{15pt}p{15pt}}
        \toprule
        \multicolumn{2}{c}{Transformers} & \multicolumn{2}{c}{MA} & FID & LPIPS & FVD & GFL & Throu\\ 
        Vanilla & Axial & Concat & Cross & $\downarrow$ &$\downarrow$ &$\downarrow$ &OPs$\downarrow$ & ghput$\uparrow$\\
		\midrule
		 $\sqrt{ }$ & & $\sqrt{ }$ & & 63.4 & 0.20 & 30.4 & 85.1 & 12.3 \\ 
		 $\sqrt{ }$ & & & $\sqrt{ }$ & 62.0 & 0.20 & 30.5 & 96.8 & 11.5 \\
         & $\sqrt{ }$ & & $\sqrt{ }$ & 62.7 & 0.20 & 31.7 & 52.7 & 37.2 \\
         \bottomrule
    \end{tabular}
    \vspace{-0.6em}
    \caption{Ablation study under deterministic video generation on CATER-GEN-v1.}
    \label{tab:ablation}
\end{table}

\section{Additional Visualizations}
\label{appendix:sec3}
\subsection{Attention Visualization in Motion Anchor}
\begin{figure*}[htbp]
    \centering
    \includegraphics[width=0.95\textwidth]{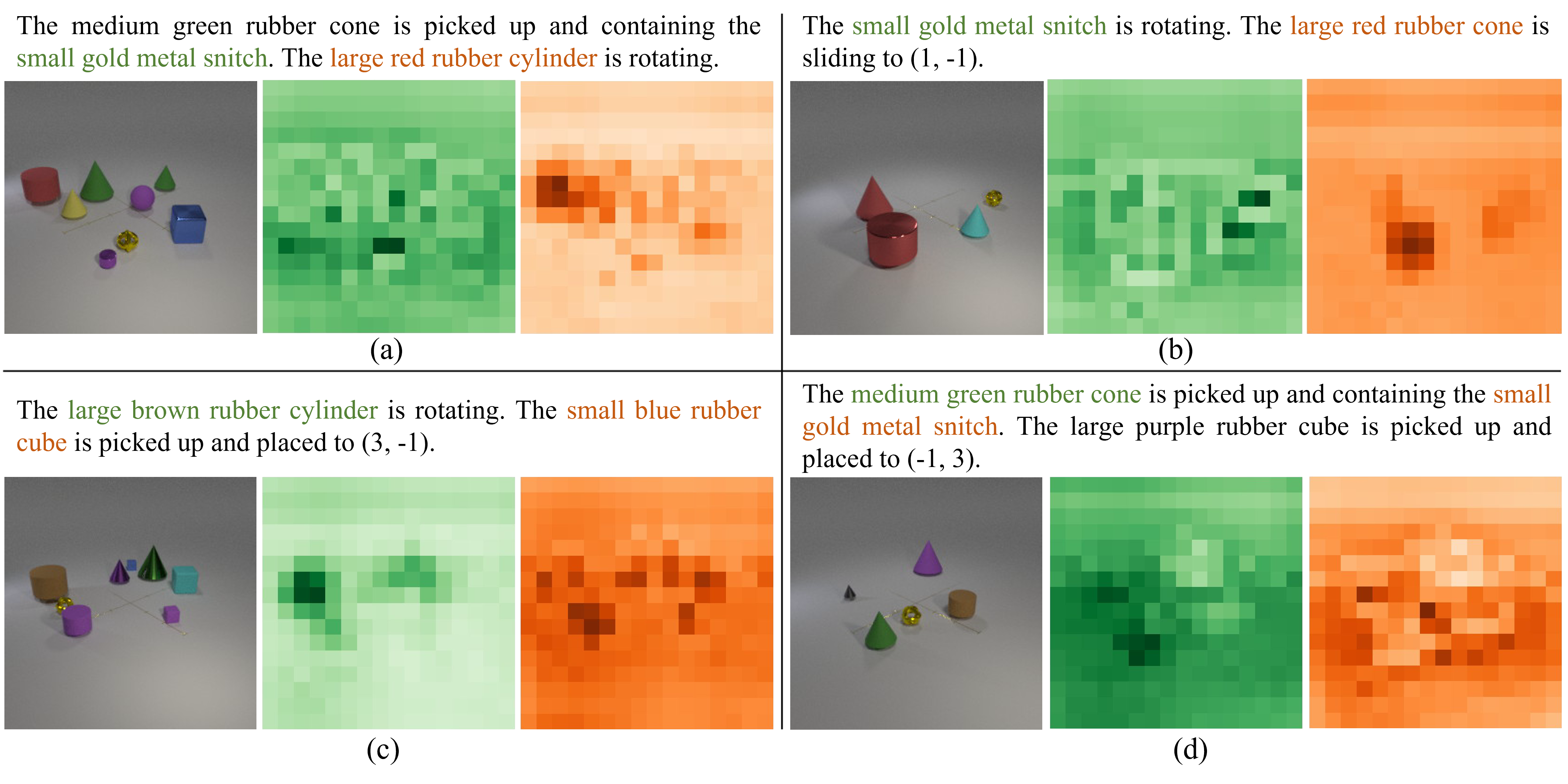}
    \caption{Visualization of attention weights from cross attention during generating motion anchor. With a group of visual token embeddings ($16\times16$) from the corresponding position in image as query, each element in attention map stands for the mean weight of 4 word embeddings of specified noun phrase marked with same color in text. The darker the color, the greater the response.}
    \label{fig:attention}
\end{figure*}

To show whether the motion anchor locates right objects and their motion, we visualize the attention map in cross-attention when generating motion anchor. Since the semantics of motion and object information in text have interacted and fused in the front text encoder, we select integral noun phrase composed of 4 attributes to show the response in image. As shown in Fig.\ref{fig:attention}, with visual token embeddings as query, we average the attention maps of different heads and show the mean attention weights of specified noun phrase (marked with same color in text). The visualizations show that the cross-attention operation is aware of multiple objects in the scene and locates the specified objects.

\begin{figure}[htbp]
    \centering
    \includegraphics[width=0.48\textwidth]{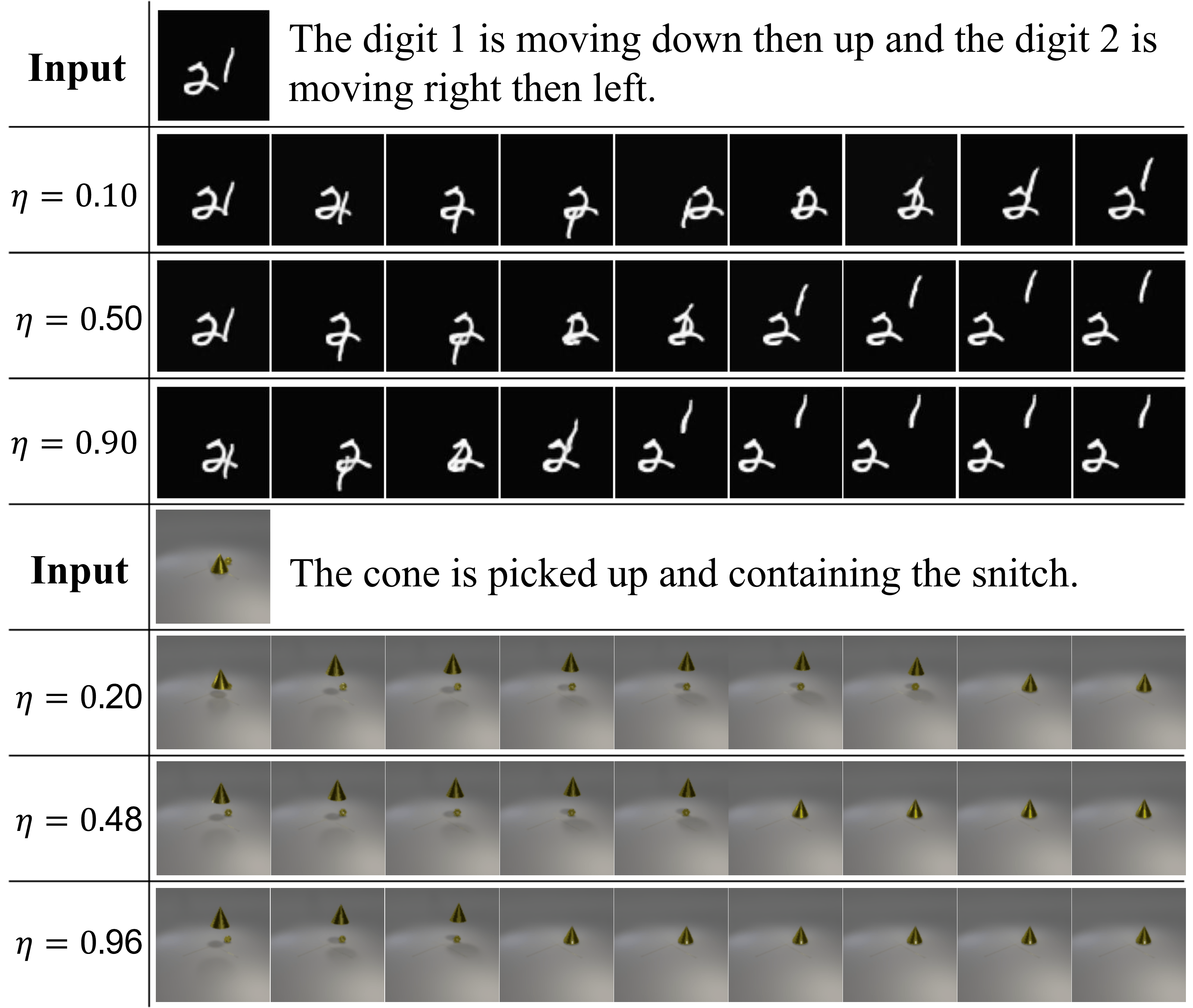}
    \caption{Generated Samples from Modified Double Moving MNIST and CATER-GEN-v1 for explicit condition evaluation. The input row is the given image and description. The left column is the input speed.}
    \label{fig:results2.1}
\end{figure}

\subsection{Visualization of Explicit Condition}
\begin{figure*}[htbp]
    \centering
    \includegraphics[width=1.0\textwidth]{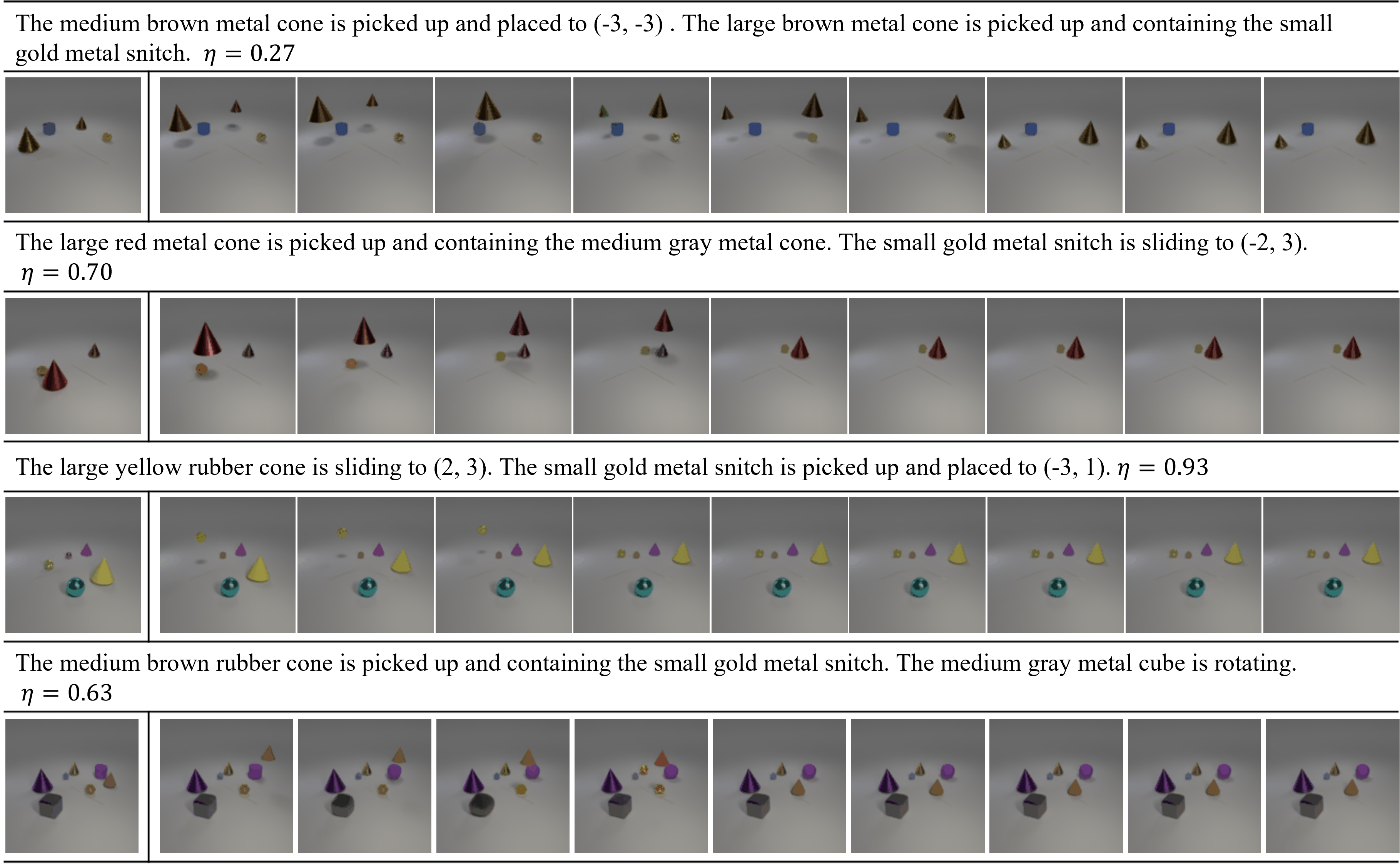}
    \caption{Generated Samples from CATER-GEN-v2 under deterministic video generation.}
    \label{fig:deterministic2}
\end{figure*}

To visualize the effect of explicit condition speed, we give the same image and description but input different speeds. Examples are shown in Fig.\ref{fig:results2.1}. Suppose each video contains 20 frames and the predefined sampling interval in training is $(1,2)$, then $\eta=0.50$, for example, stands for corresponding sampling interval with 1.5. By giving different $\eta$, it can be found that the model correctly generates videos with corresponding speed. More generated videos from CATER-GEN-v2 are shown in Fig.\ref{fig:deterministic2}.

\begin{figure*}[htbp]
    \centering
    \includegraphics[width=0.98\textwidth]{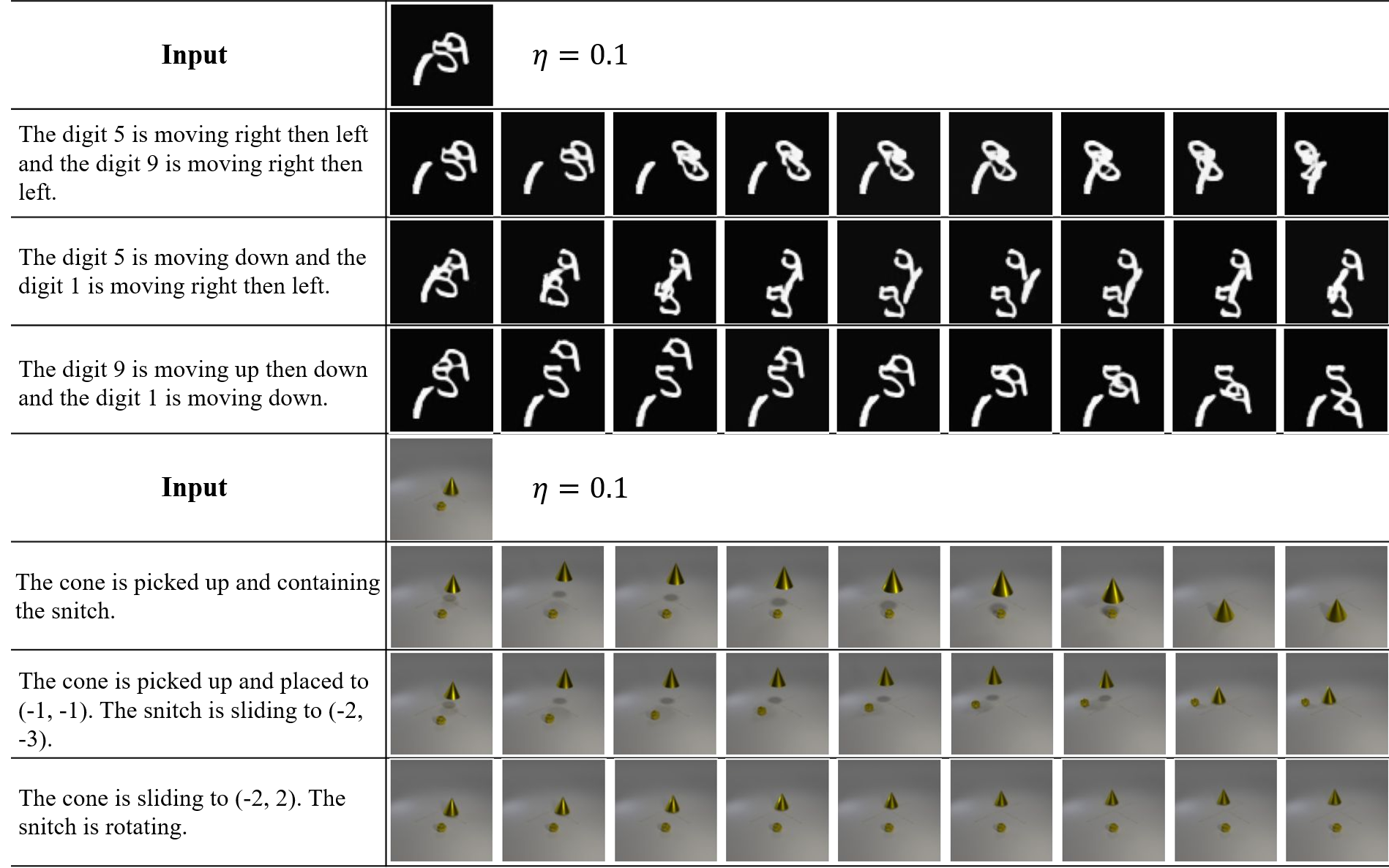}
    \caption{Generated Samples from Modified Double Moving MNIST and CATER-GEN-v1 for composability evaluation.}
    \label{fig:results3.3}
\end{figure*}

\subsection{Visualization of Composability}
More compositional video generation results from Modified Double Moving MNIST and CATER-GEN-v1 are shown in Fig.\ref{fig:results3.3}. Given an image and a fixed speed, three descriptions are input separately to specify different objects and actions. Results show correct concordance with text both on moving targets and actions. 

\subsection{Visualization of Implicit Randomness}
\begin{figure*}[htbp]
    \centering
    \includegraphics[width=0.98\textwidth]{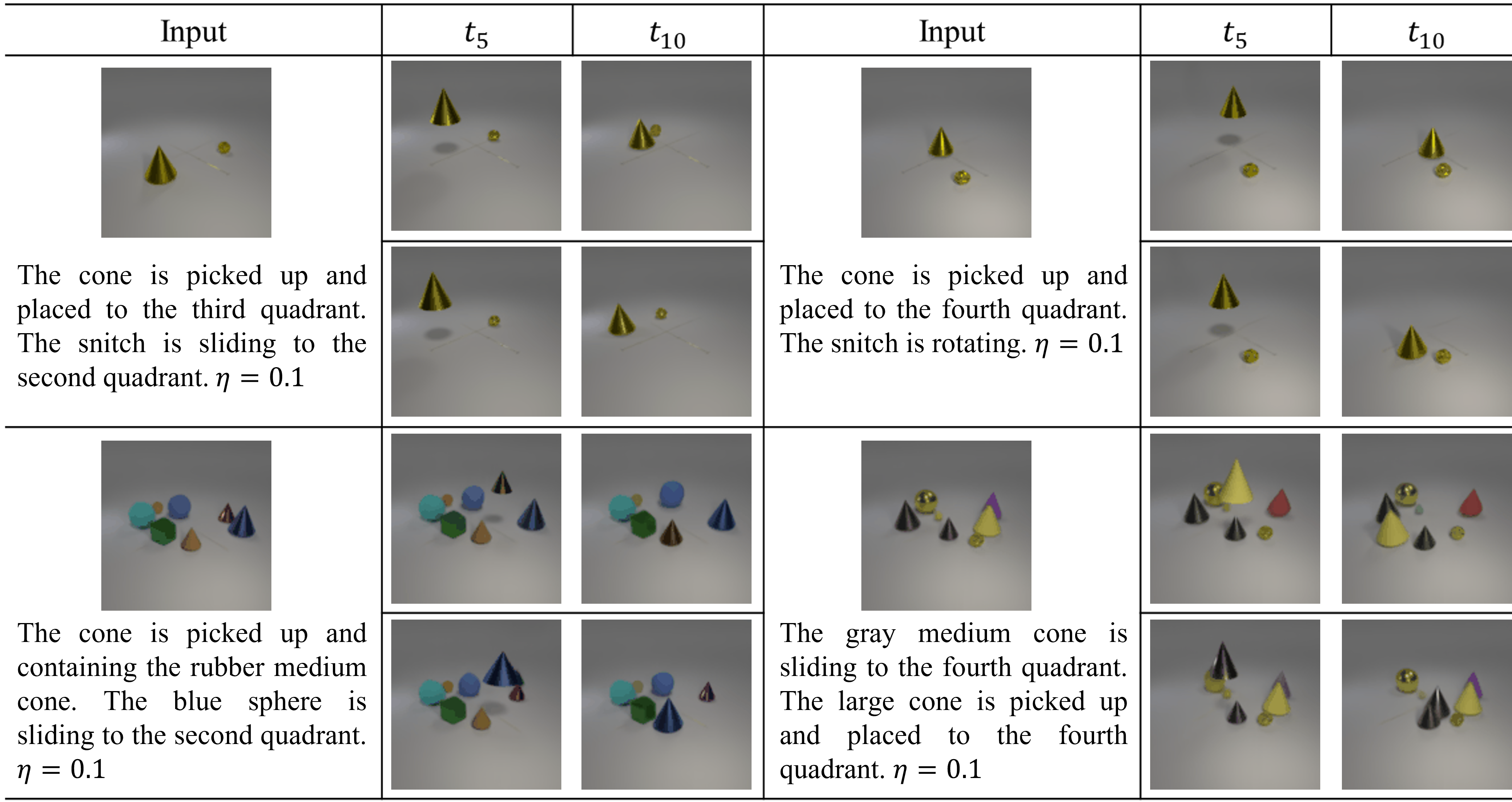}
    \caption{Generated Samples from CATER-GEN-v1 and CATER-GEN-v2 for diverse video generation.}
    \label{fig:results4.3}
\end{figure*}

We also show diverse generated videos in Fig.\ref{fig:results4.3} with ambiguous text as input. The 5$th$ frames and 10$th$ frames from two generated video with the same input are shown. It can be found that, even given a difficult image and a complicated caption, our method can model the implied randomness (including final position or action subject) and generate diverse and relatively satisfactory results. However, when the appearance of image is too complicated, there may be deformation and clipping problems (like the last row). Besides, since the VQ-VAE is trained on frame-level, there are some objects in reconstructed videos have color change. That will result in color inconsistency of generated videos (like the third row on the right).

\begin{figure*}[htbp]
    \centering
    \includegraphics[width=0.98\textwidth]{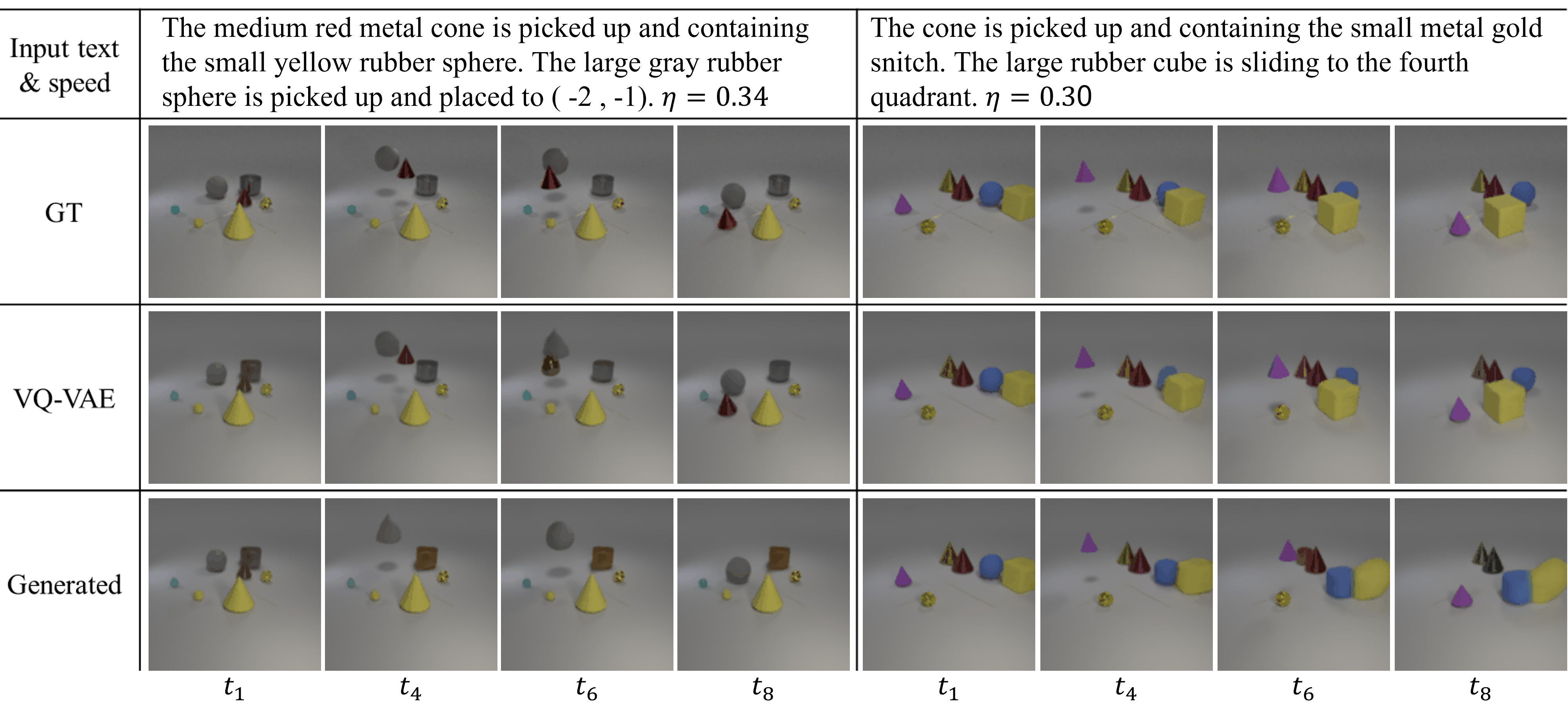}
    \caption{Failure Cases from CATER-GEN-v2. Four rows from top to bottom are input text and speed, the reference video, the reconstructed video from VQ-VAE only, and the generated video from MEGA, respectively. For each video, the $1st$, $4th$, $6th$, and $8th$ frames are shown.}
    \label{fig:failures}
\end{figure*}

\begin{figure*}[htbp]
    \centering
    \includegraphics[width=0.92\textwidth]{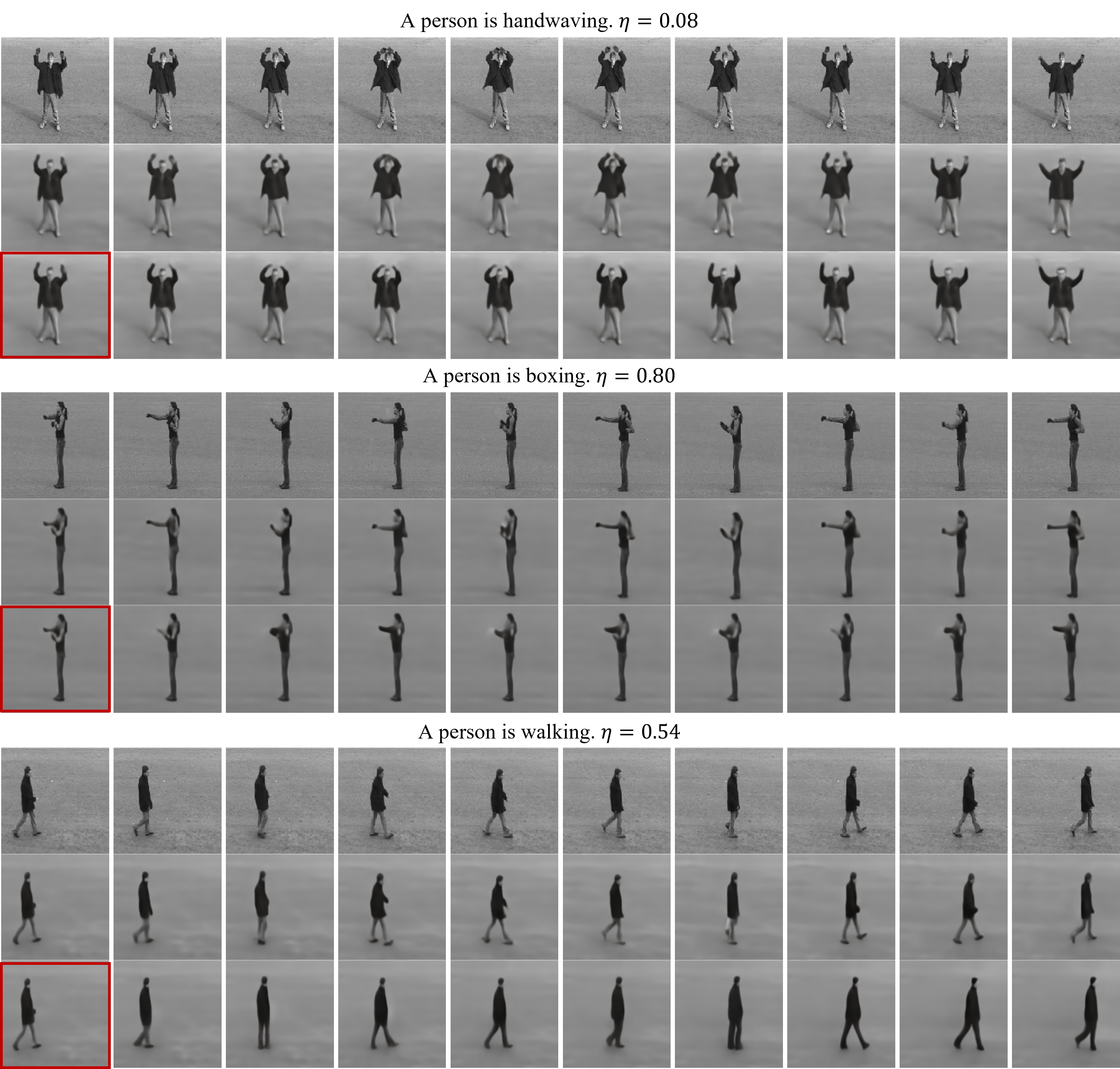}
    \caption{Generated Samples from KTH. Three rows from top to bottom of each example are the reference video, the reconstructed video from VQ-VAE only, and the generated video from MEGA, respectively. The red box represents the first given image when generating video.}
    \label{fig:results3.1}
\end{figure*}
\subsection{Failure Cases}
We also show some failure examples in Fig.\ref{fig:failures}. The first example shows the effect of VQ-VAE. In our method, VQ-VAE is responsible for image tokenizer before generation (from $128 \times 128$ to $16 \times 16$) and reconstruction after generation. Therefore, the performance of VQ-VAE will directly affect video generator. Because of the large down-sampling ratio, VQ-VAE may lose fine-grained texture and result in wrong shape of reconstructed object (the gray sphere in the $6th$ frame of reconstruction-only video), further leading to distortion of generated video (the gray sphere in the $4th$ frame of generated video). 

On the other hand, since MEGA generates video tokens at latent space with spatial size $16 \times 16$, the small resolution makes it hard to split two objects with overlap (like the blue sphere and yellow cube in the second example). This may also cause distortion when two objects intersect during moving. 

\subsection{Visualization of Realistic Video Generation}
Except for synthetic videos, we also wonder whether our model can generate realistic videos. However, existing paired video-text datasets contain high uncertainty and much noise and like scene change or the emergence of irrelevant objects. The texts are also not fine-grained enough, making video generation hard to control. Therefore, we evaluate our method on KTH \cite{schuldt2004recognizing} which is a relatively clean action recognition dataset. The KTH dataset contains 2391 video clips of six human actions performed by 25 people in four different scenarios. We use the original train-test split. The text is formed like ``A person is $[action\_label]$." The reference video, reconstructed video of VQ-VAE only and generated video of MAGE are shown in Fig.\ref{fig:results3.1}. Compared to reference videos, the reconstructed videos are blurrier and lost details due to the limitation of VQ-VAE. Even though, generated videos do not have much degradation in quality compared to reconstructed videos. The speed of generated videos is also consistent with input. However, for action like ``walking", there may be no person in the first given image, resulting in that the model tends to learn an ``average person" (like a body with black color) in the training stage. This may cause the body to gradually turn black during generation. It also shows that generating unseen objects is still a challenge.

\section{Limitations}
Although our method can generate controllable and diverse videos, there are failure cases that reflect some limitations of MAGE. Besides, TI2V task also faces demands in evaluation metrics and datasets. We summarize those limitations here.

\begin{itemize} 
    \item \textbf{VQ-VAE based Architecture:} Despite VQ-VAE greatly reduces the data volume and facilitates the training of video generator, the reconstruction performance of VQ-VAE will directly affect prediction accuracy of subsequent video generator. Especially under a large down-sampling ratio, VQ-VAE may lose fine-grained texture, leading to distortion of object. Besides, as VQ-VAE is trained on image-level, the temporal consistency is not guaranteed, which may result in inconsistency of predicted videos. Therefore, higher performance VQ-VAE and proper down-sampling ratio should be considered to help model generate more consistent and high-resolution videos.
    \item \textbf{Evaluation Metrics:} Evaluating the quality of generated videos is challenging for TI2V task, especially for ambiguous text descriptions. Since the ``correct" video may not be unique and accessible, existing similarity metrics (e.g. PSNR, LPIPS, FVD) measuring the semantic consistency between generated video with one of ``correct" videos are not accurate. Meanwhile, diversity metrics (e.g. DIV VGG/I3D) can only reflect the variation between generated videos no matter whether they are consistent with text descriptions. Thus, appropriate metrics to evaluate both accuracy and diversity are needed.
    \item \textbf{Realistic Video Generation:} The difficulties for realistic video generation lie in not only the great uncertainty in realistic videos, but also the lack of appropriate datasets. Most of existing video-text paired datasets are composed of coarse-grained text descriptions, making it harder to generate coherent motion given the first image. Generating realistic and open-domain videos is still a major challenge of TI2V task.
\end{itemize}

\end{document}